\documentclass[entropy,article,accept,moreauthors,pdftex]{Definitions/mdpi} 
\usepackage[labelformat=simple]{subcaption}

\DeclareCaptionLabelFormat{subcaptionlabel}{\normalfont(\textbf{#2}\normalfont)}
\usepackage{caption}

\usepackage[utf8]{inputenc} %
\usepackage[T1]{fontenc}    %
\usepackage{hyperref}       %
\usepackage{url}            %
\usepackage{booktabs}       %
\usepackage{amsfonts}       %
\usepackage{nicefrac}       %
\usepackage{microtype}      %
\usepackage{xcolor}         %
\usepackage[framemethod=TikZ]{mdframed}

\usepackage{amsmath,amsfonts,bm}

\def\eqref#1{equation~\ref{#1}}

\def\1{\bm{1}}

\DeclareMathAlphabet{\mathsfit}{\encodingdefault}{\sfdefault}{m}{sl}
\SetMathAlphabet{\mathsfit}{bold}{\encodingdefault}{\sfdefault}{bx}{n}

\newcommand{\R}{\mathbb{R}}

\usepackage{url}
\usepackage{microtype}
\usepackage{graphicx}
\usepackage{booktabs} %
\usepackage{bbm}
\usepackage{wrapfig} 
\usepackage{microtype}
\usepackage{booktabs} %
\usepackage{wrapfig}
\usepackage{algorithm}
\usepackage[noend]{algorithmic}
\usepackage{tabularx}
\usepackage{amsmath}
\usepackage{amssymb}
\usepackage{mathrsfs}
\usepackage{amsthm}
\usepackage{stmaryrd}
\usepackage{mathtools}
\usepackage{bm}
\usepackage{dblfloatfix}
\usepackage{enumitem}
\usepackage[thinc]{esdiff}
\usepackage{bigints}
\usepackage{mdwlist}
\usepackage{thmtools, thm-restate}
\usepackage{pgfplots}
\usepackage{subcaption}
\usepackage{caption}
\usepackage{subcaption}
\declaretheorem[numbered=no]{principle}
\usepackage{amsmath}
\usepackage{centernot}

\newcommand{\Xin}{\mathcal{X}_{in}}
\newcommand{\Xout}{\mathcal{X}_{out}}
\newcommand{\abs}[1]{\left\lvert #1 \right\rvert}
\newcommand{\norm}[1]{\left\lVert #1 \right\rVert}

\firstpage{1} 
\pubvolume{23}
\issuenum{1}
\articlenumber{1690}
\doinum{10.3390/e23121690}
\pubyear{2021}
\copyrightyear{2021}
\externaleditor{{Academic Editor}: Eric Nalisnick, Boris Ryabko} 
\datereceived{30 September 2021} 
\dateaccepted{13 December 2021} 
\datepublished{16 December 2021} 
\hreflink{https://doi.org/10.3390/
\linebreak
e23121690
} %
\Title{Perfect Density Models Cannot Guarantee Anomaly Detection}

\TitleCitation{Perfect Density Models Cannot Guarantee Anomaly Detection}

\Author{{Charline Le Lan} $^{1,2,}$* and Laurent Dinh $^{2}$}

\AuthorNames{Charline Le Lan and Laurent Dinh}

\AuthorCitation{Le Lan, C.; Dinh, L.}

\address{%
$^{1}$ \quad Department of Statistics, University of Oxford, Oxford OX1 3LB, UK\\
$^{2}$ \quad {Google Research, Montreal H3B 2Y5, CA}}

\corres{Correspondence: charline.lelan@stats.ox.ac.uk}

\abstract{Thanks to the tractability of their likelihood, several deep generative models show
promise for seemingly straightforward but important applications like anomaly
detection, uncertainty estimation, and active learning. However, the likelihood
values empirically attributed to anomalies conflict with the expectations these
proposed applications suggest. In this paper, we take a closer look at the
behavior of distribution densities through the lens of reparametrization and show that these quantities carry less
meaningful information than previously thought, beyond estimation issues
or the curse of dimensionality. We conclude that the use of these likelihoods for
anomaly detection relies on strong and implicit hypotheses, and
highlight the necessity of explicitly formulating these assumptions for reliable
anomaly detection.}


\begin{document}

\section{Introduction}
Several machine learning methods aim at extrapolating a behavior observed on
training data in order to produce predictions on new observations. However, every so
often, such extrapolation can result in wrong outputs, especially on points that
we would consider infrequent with respect to the training distribution. 
Faced with unusual situations, whether
adversarial~\citep{szegedy2013intriguing,carlini2017adversarial} or just
rare~\citep{hendrycks2018benchmarking},
a desirable behavior from a machine learning system would be to flag these
{\em outliers} so that the user can assess if the result is reliable
and gather more information if it should be necessary~\citep{zhao2019curiosity,fu2017ex2}.
This can be critical for applications like
medical decision making~\citep{lee2018simple} or autonomous vehicle
navigation~\citep{filos2020can}, where such outliers \mbox{are ubiquitous}.

\looseness=-1 What are the situations that are deemed unusual? Defining these
{\em anomalies}~\citep{grubbs1969procedures, barnett1984outliers, hodge2004survey,pimentel2014review, ruff2021unifying}
manually can be laborious if not impossible, and so generally
applicable, automated methods are preferable. In that regard, the
framework of {\em probabilistic reasoning} has been an appealing formalism
because a natural candidate for outliers are situations that are
{\em improbable}.
Since the true {\em probability distribution density} $p^*_X$ of the data is
mostly not provided, one would instead use an estimator $p^{(\theta)}_X$ from
this data to assess the regularity of a point.

Density estimation has been a particularly challenging task on
high-dimensional problems.
However, recent advances in {\em deep probabilistic models},
including variational
auto-encoders~\citep{kingma2013auto,rezende2014stochastic,vahdat2020nvae}, deep
autoregressive
models~\citep{uria2014deep,oord2016pixel,van2016conditional},
and flow-based generative
models~\citep{dinh2014nice,dinh2016density,kingma2018glow,ho2019flow++,kobyzev2020normalizing,papamakarios2019normalizing}, have shown promise
for density estimation, which has the potential to enable accurate
{\em density-based methods}~\citep{bishop1994novelty} for anomaly detection.

Yet, several works have observed that a significant gap persists between the
potential of density-based anomaly detection and empirical results.
For instance \cite{choi2018waic,nalisnick2018deep,hendrycks2018deep} noticed that generative models trained on a benchmark
dataset~(e.g., CIFAR-10, \cite{krizhevsky2009learning}) and tested on
another~(e.g., SVHN, \cite{netzer2011reading}) are not able to identify the
latter as  an outlier with current methods. Different
hypotheses have been formulated to explain that discrepancy, ranging from the
{\em curse of dimensionality}~\citep{nalisnick2019detecting} to a significant
{\em mismatch between $p^{(\theta)}_X$ and $p^*_X$}~\citep{choi2018waic, just2019deep,
Fetaya2020Understanding, kirichenko2020normalizing, zhang2020hybrid, wang2020further}.

\begin{figure}[H]
\widefigure
  \begin{subfigure}[t]{0.3\textwidth}
  \begin{tikzpicture}
    \node (img)  {\includegraphics[width=.9\textwidth]{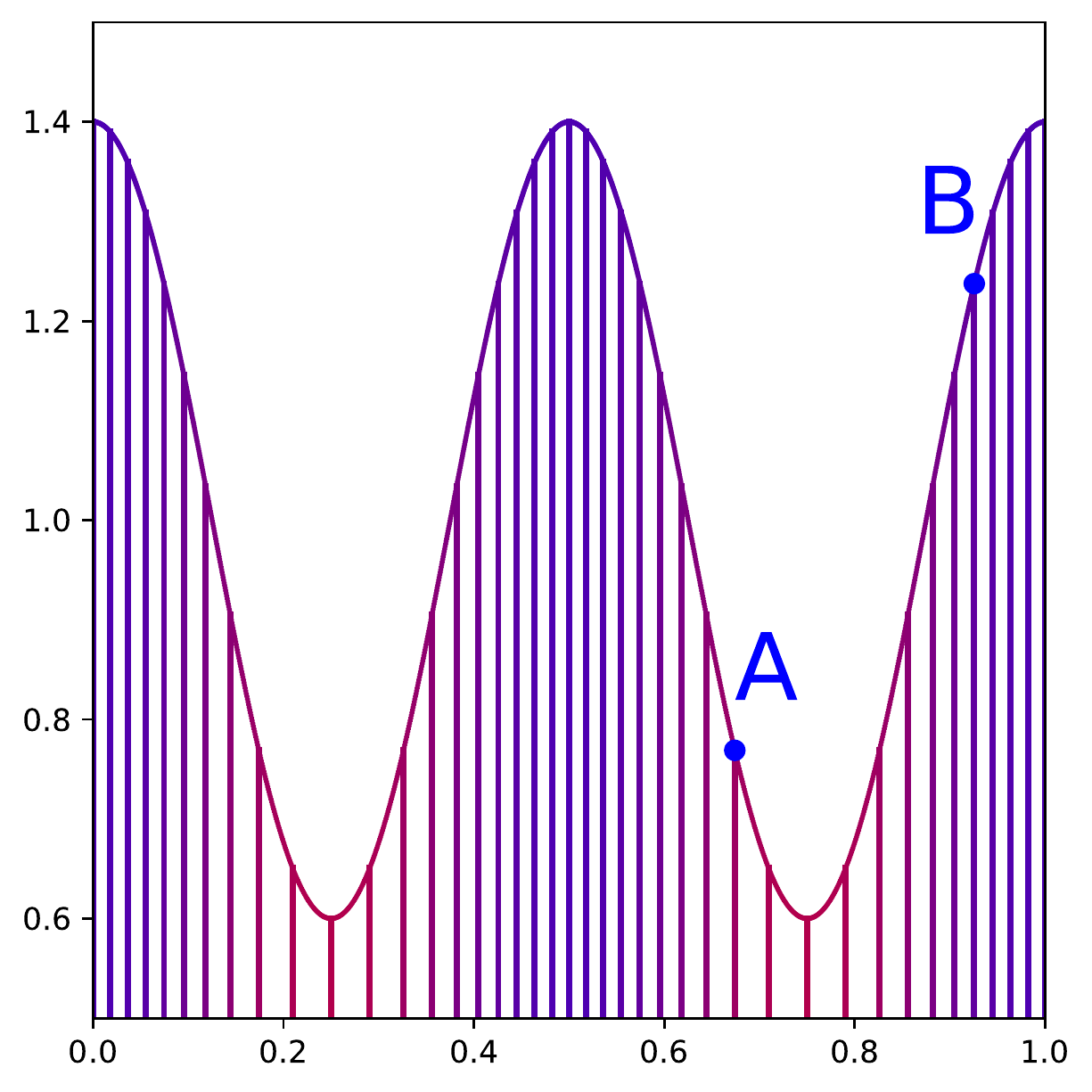}};
    \node[below of=img, node distance=2.6cm, yshift=0cm] {\small $x$};
    \node[left of=img, node distance=2.6cm, rotate=90, anchor=center] {\small $p^*_X(x)$};
  \end{tikzpicture}
  \caption{}
  \label{fig:pdf_1_cov}
  \end{subfigure}
  \begin{subfigure}[t]{0.3\textwidth}

  \begin{tikzpicture}
    \node (img)  {\includegraphics[width=.9\textwidth]{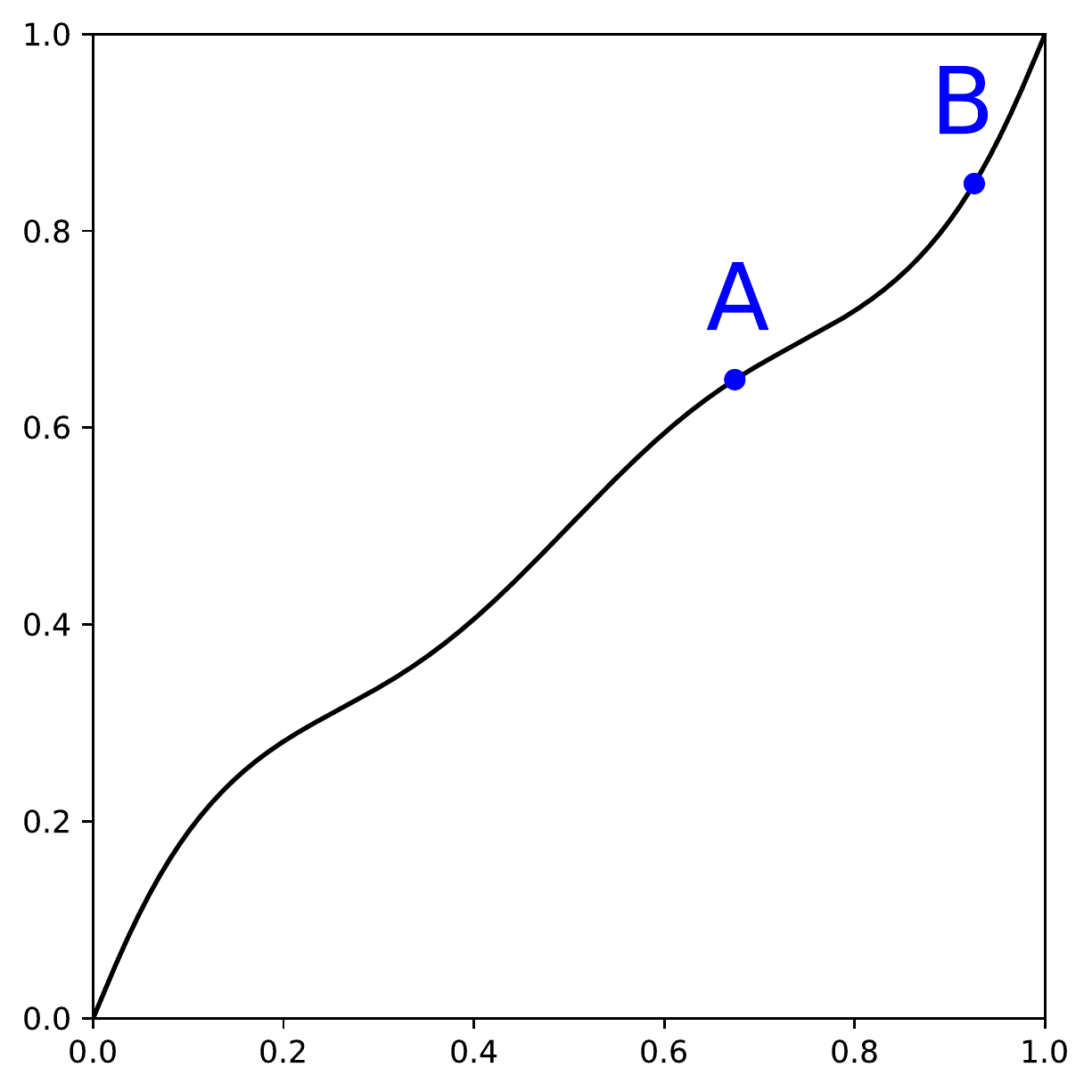}};
    \node[below of=img, node distance=2.6cm, yshift=0cm] {\small $x$};
    \node[left of=img, node distance=2.6cm, rotate=90, anchor=center] {\small $z = f(x)$};
  \end{tikzpicture}

  \caption{}
  \label{fig:pdf_1_to_pdf_3_cov}
  \end{subfigure}
  \begin{subfigure}[t]{0.3\textwidth}
  \begin{tikzpicture}
    \node (img)  {\includegraphics[width=.9\textwidth]{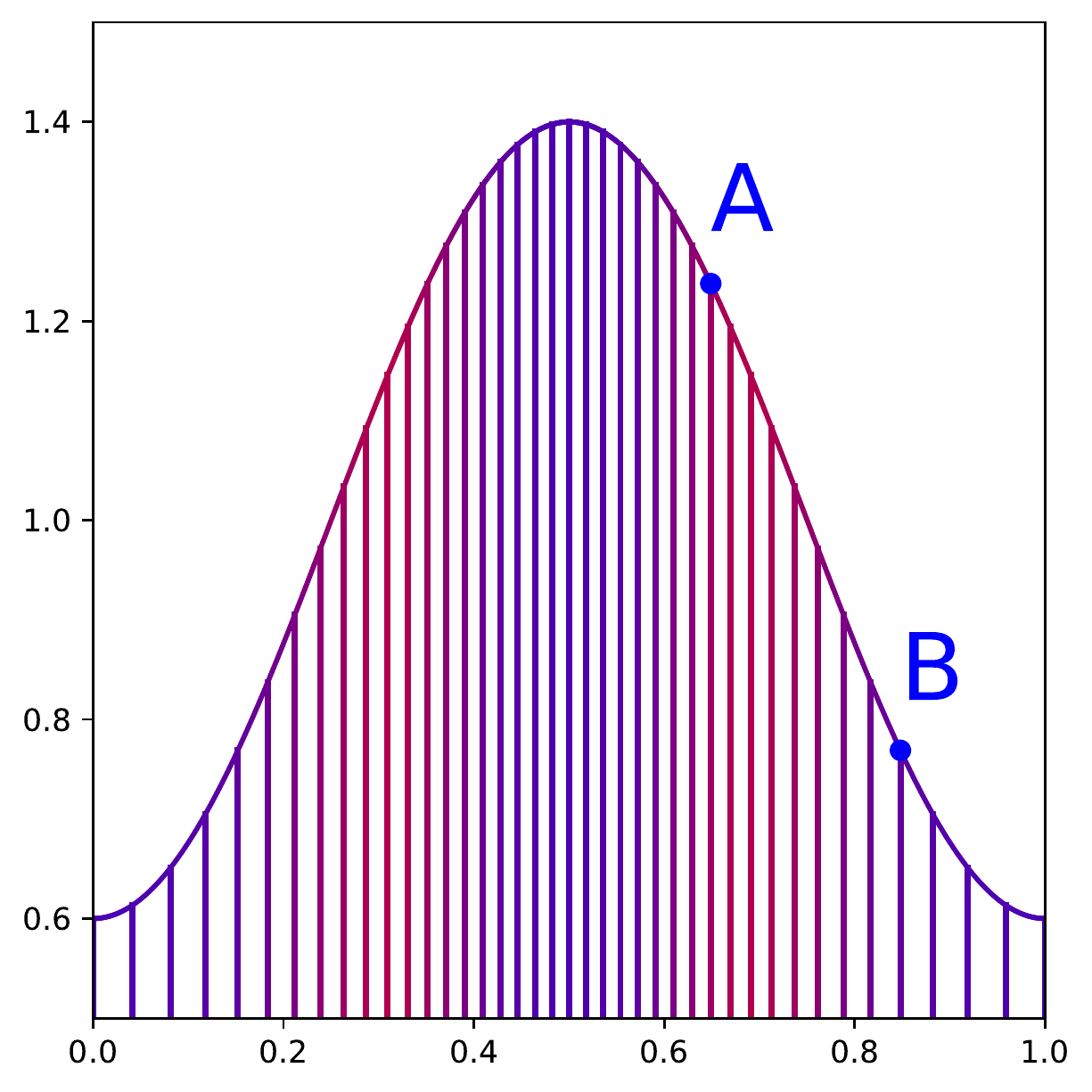}};
    \node[below of=img, node distance=2.6cm, yshift=0cm] {\small $z$};
    \node[left of=img, node distance=2.6cm, rotate=90, anchor=center] {\small $p^*_{Z}(z)$};
  \end{tikzpicture}
  \caption{}
  \label{fig:pdf_3_cov}
  \end{subfigure}
  \vspace{6pt}
  \caption{
  An invertible change of representation can affect the relative density between two points A and B, which has been interpreted as their relative regularity. (\textbf{a}) An example of a distribution density $p^*_X$. (\textbf{b}) Example of an invertible function $f$ from $[0, 1]$ to $[0, 1]$. (\textbf{c}) {Resulting density} $p^*_{Z}$ as a function of the new axis $z = f(x)$. In (\textbf{a},\textbf{c}) points with high original density $p^*_X(x)$ are in blue and red for low original density.}
  \label{fig:change-of-variables}
\end{figure}
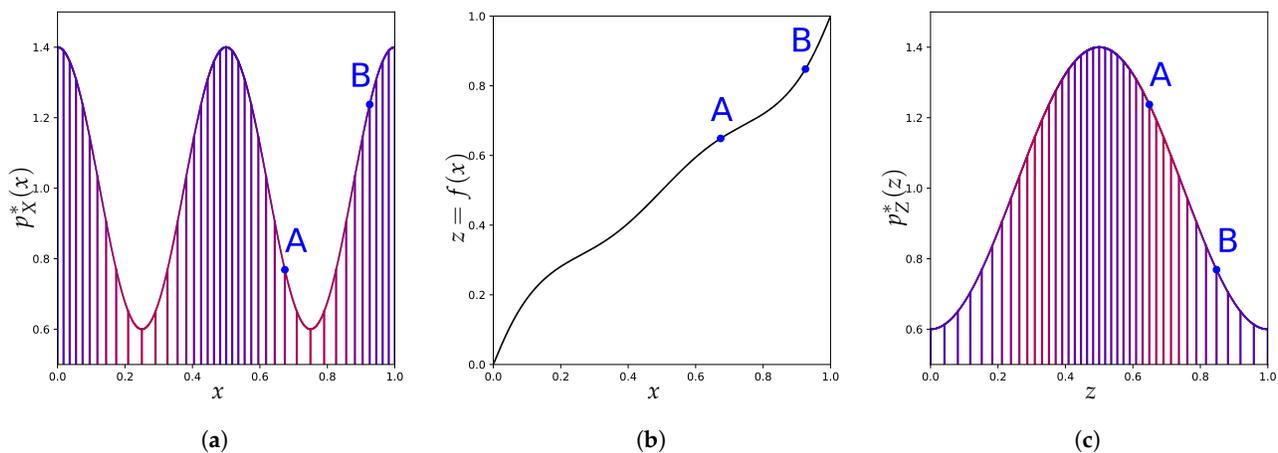

In this work, we propose a new perspective on this discrepancy and challenge
the expectation that density estimation should always enable anomaly detection.
We show that the aforementioned discrepancy persists even with perfect density
models, and therefore goes beyond issues of estimation, approximation,
or optimization errors~\citep{bottou2008tradeoffs}.
We highlight that this issue is pervasive as it occurs even in low-dimensional
settings and for a variety of density-based methods for anomaly detection.
Focusing on the continuous input case, we make the following contributions:
\begin{itemize}[leftmargin=*]
\item Similar to classification, we propose in Section \ref{sec:idea} a principle of invariance to formalize the underlying assumptions behind the current practice of (deep) density-based methods.
\item We use the well-known change of variables formula for probability density to show in Section \ref{sec:chofvar} how these density-based methods are not invariant to reparametrization (see Figure \ref{fig:change-of-variables}) and contradict this principle. We demonstrate the extent of the issues with current practices by building adversarial cases, even under strong distributional constraints. %
\item  Given the resulting tension between the use of these anomaly detection methods and their lack of invariance, we focus in Section \ref{sec:prior} on the importance of explicitly incorporating prior knowledge into (density-based) anomaly detection methods as a more promising avenue to reconcile this tension.
\end{itemize}

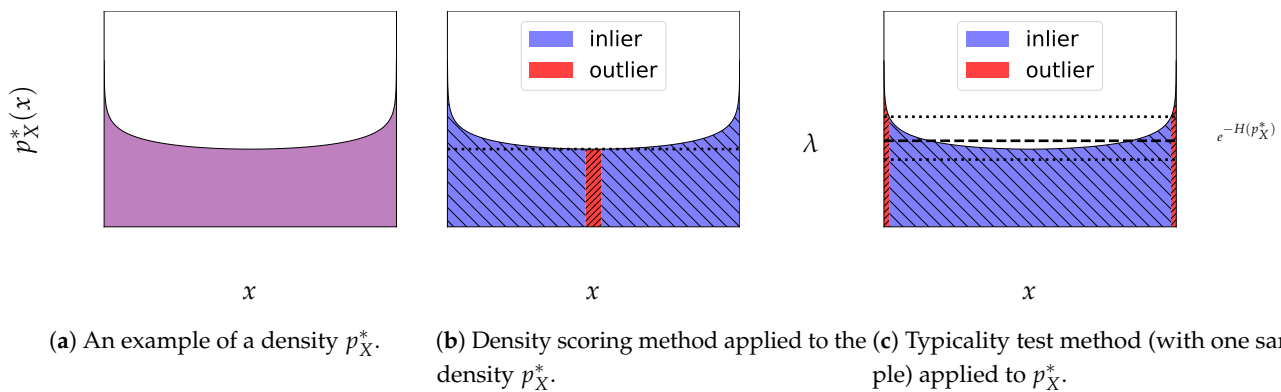
\begin{figure}[H]
\widefigure
  \begin{subfigure}[t]{0.3\textwidth}
  \begin{tikzpicture}
    \node (img)  {\includegraphics[scale=0.2]{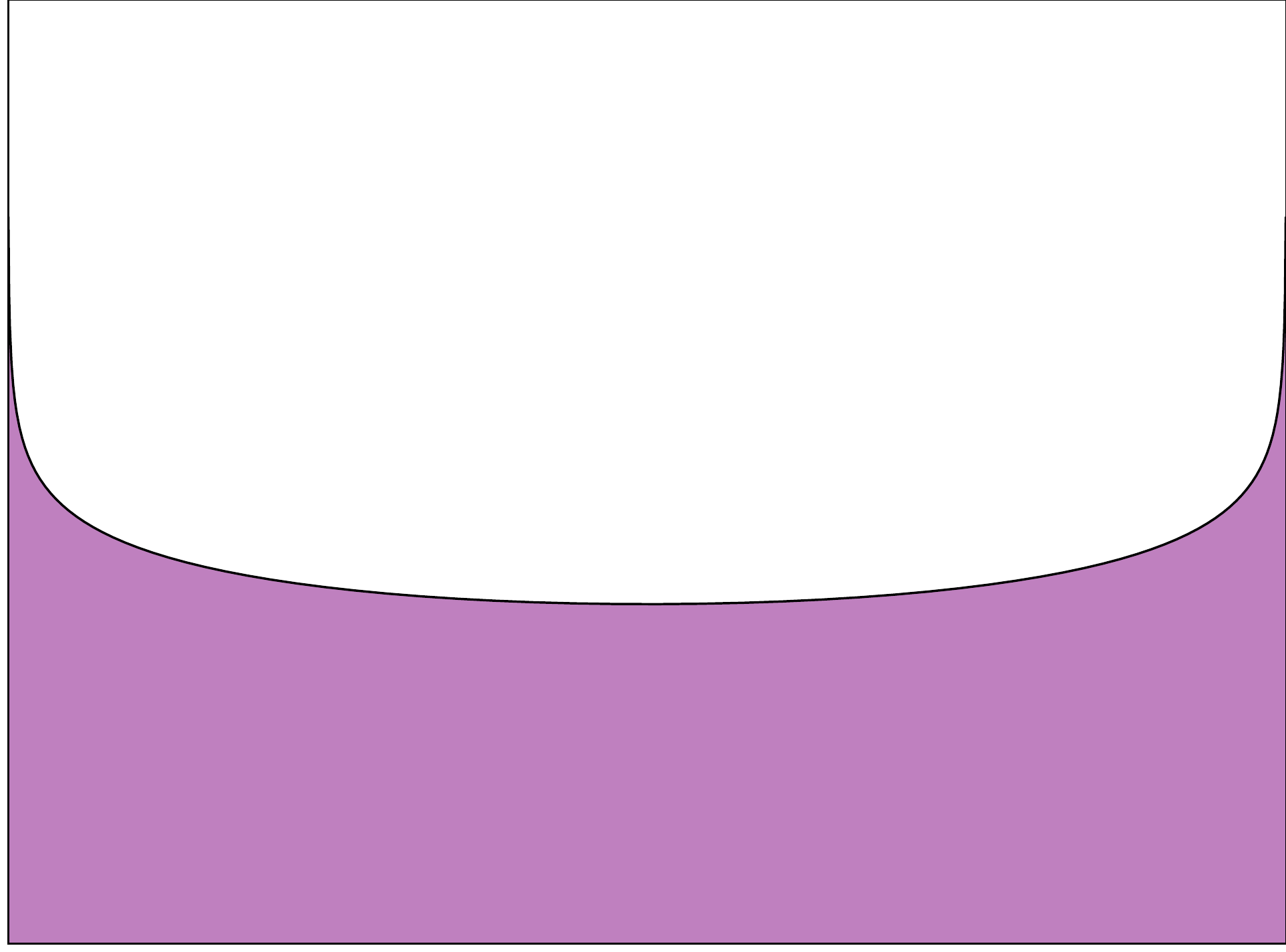}};
    \node[below of=img, node distance=2.3cm, yshift=0cm] {$x$};
    \node[left of=img, node distance=3.cm, rotate=90, anchor=center] {$p^*_X(x)$};
  \end{tikzpicture}
  \caption{An example of a density $p^*_X$.}
  \label{fig:example-dist}
  \end{subfigure}
  \begin{subfigure}[t]{0.3\textwidth}
  \begin{tikzpicture}
    \node (img)  {\includegraphics[scale=0.2]{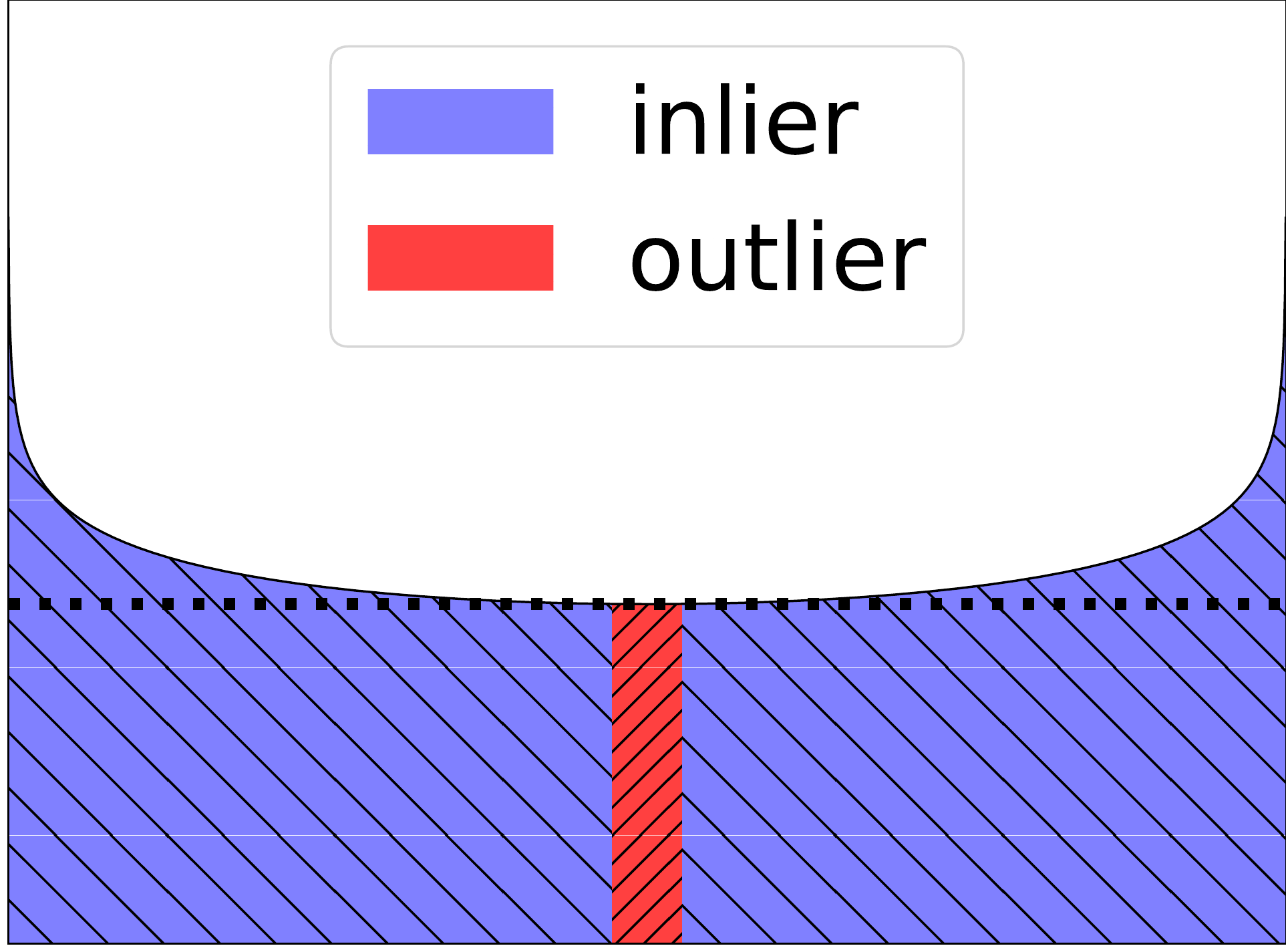}};
    \node[below of=img, node distance=2.3cm, yshift=0cm] {$x$};
    \node(capt) [right of=img, node distance=2.9cm, anchor=center] {};
    \node[below of=capt, node distance=0.3cm, anchor=center] {$\lambda$};
  \end{tikzpicture}
  \caption{Density scoring method applied to the density $p^*_X$.}
  \label{fig:example-ds}
  \end{subfigure}
  \begin{subfigure}[t]{0.3\textwidth}
  \begin{tikzpicture}
    \node (img)  {\includegraphics[scale=0.2]{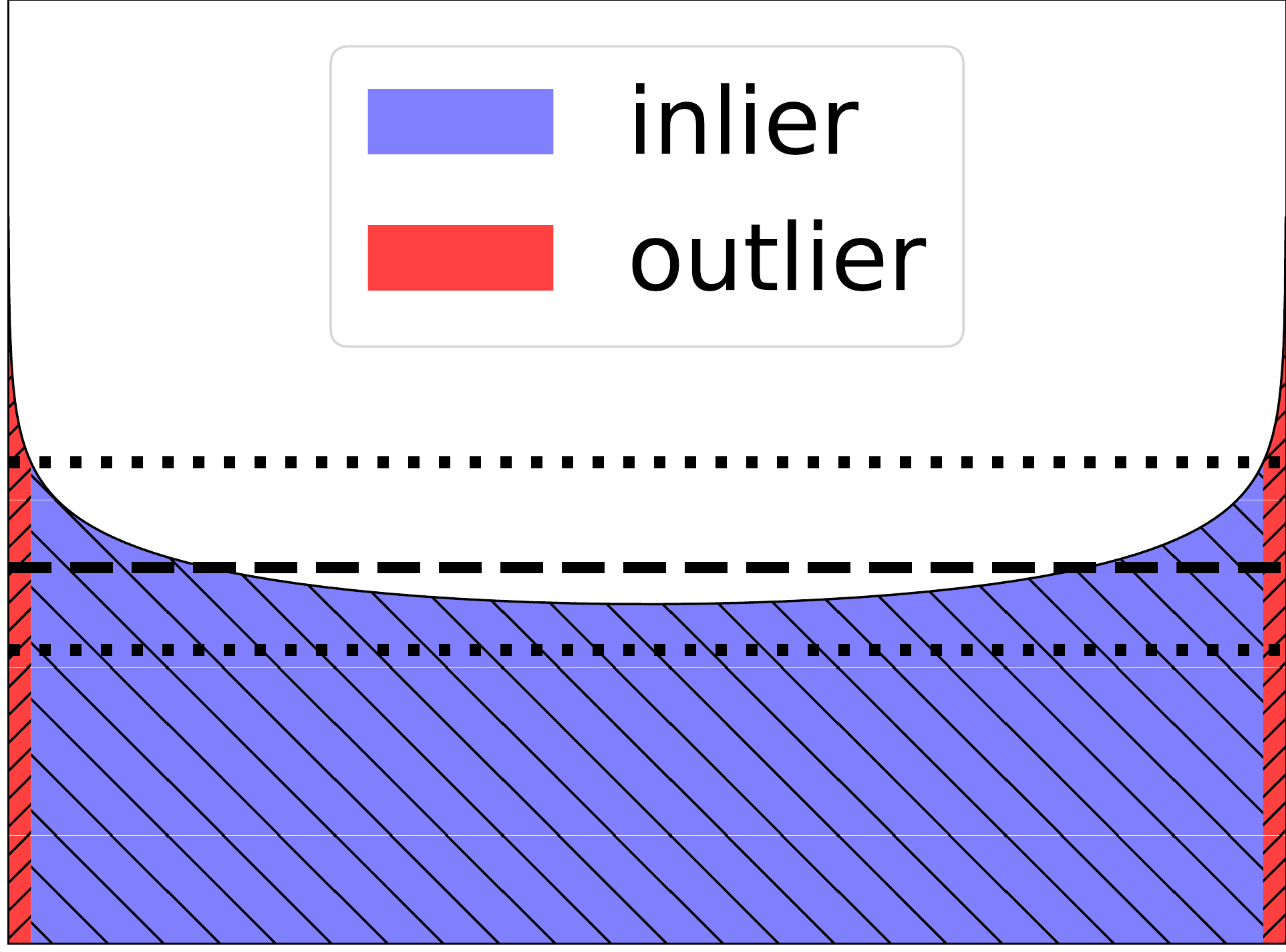}};
    \node[below of=img, node distance=2.3cm, yshift=0cm] {$x$};
    \node(capt) [right of=img, node distance=2.9cm, anchor=center] {};
    \node[below of=capt, node distance=0.15cm, anchor=center] {\tiny $e^{-H(p^*_X)}$};
  \end{tikzpicture}
  \caption{Typicality test method (with one sample) applied to $p^*_X$. }
  \label{fig:example-tt}
  \end{subfigure}
    \vspace{12pt}
  \caption{Illustration of different density-based methods applied to a particular one-dimensional distribution $p^*_X$. Outliers are in
  red and inliers are in blue. The thresholds are picked so that inliers
  include $95\%$ of the mass.
  In (\textbf{b}), inliers are
  considered as the points with density above the threshold $\lambda >0$
  while in (\textbf{c}), they are the points whose log-density
  are in the $\epsilon$-interval around the negentropy $-H(p^*_X)$.
  }
  \label{fig:density-methods}
\end{figure}

\vspace{-6pt}
\end{paracol}
\nointerlineskip
\vspace{-12pt}
\begin{paracol}{2}
\switchcolumn

\section{Density-Based Anomaly Detection}
\label{sec:bg}
In this section, we present existing density-based anomaly detection approaches that are central to our analysis. Seen as methods without explicit prior knowledge, they aim at unambiguously defining outliers and inliers.
\subsection{Unsupervised Anomaly Detection: Problem Statement}
\label{sec:pb}
Unsupervised anomaly detection is a classification problem~\citep{moya1993one,
scholkopf2001estimating, steinwart2005classification},
where one aims at distinguishing between regular
points ({\em inliers}) and irregular points ({\em outliers}).
However, as opposed to the usual
classification task, labels distinguishing inliers and outliers are
not provided for training, if outliers are even provided at all. Given an
input space $\mathcal{X} \subseteq \mathbb{R}^D$, the task can be
summarized as partitioning this space between the subset of outliers $\Xout$
and the subset of inliers $\Xin$, i.e., $\Xout \cup \Xin = \mathcal{X}$ and
$\Xout \cap \Xin = \varnothing$. When the training data is distributed
according to the probability measure $P^*_X$ (with density $p^*_X$, that we assume in the rest of the paper to be such that $\forall x \in \mathcal{X}, p^*_X(x) > 0$) one would
usually pick the set of regular points $\Xin$ such that this set contains the
majority (but not all) of the mass (e.g., $95\%$) of this distribution \citep{scholkopf2001estimating}, i.e., $P^*_X(\Xin) = 1- \alpha \in \left(\frac{1}{2}, 1\right)$. However, for any
given $\alpha$, there exists in theory an infinity of corresponding partitions into
$\Xin$ and $\Xout$ (see Figure \ref{fig:partition-gaussian}). How are
these partitions picked to match our intuition of inliers and
outliers? In particular, how can we {\em learn} from data to discriminate between inliers and outliers (without of course predefining them)? We will focus in this paper on recently used methods based on
probability density.

\begin{figure}[H]

  \begin{tikzpicture}
    \node(img){\includegraphics[width=.54\textwidth]{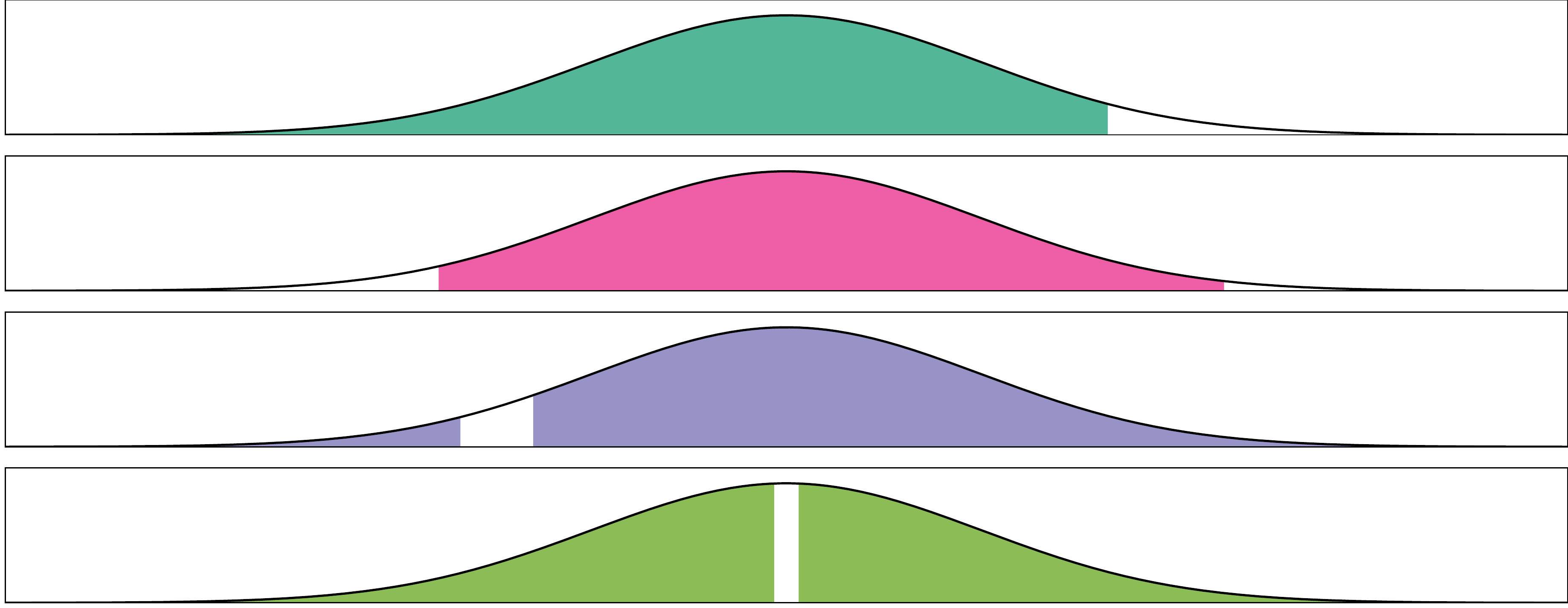}};
    \node[below of=img, node distance=2.4cm, yshift=0cm] {$x$};
    \node[left of=img, node distance=5.7cm, rotate=90, anchor=center] {$p^*_X(x)$};
  \end{tikzpicture}
  \caption{There is an infinite number of ways to partition a distribution in
  two subsets, $\Xin$ and $\Xout$ such that $P^*_X(\Xin) = 0.95$. Here, we show
  several choices for a standard Gaussian $p^*_X = \mathcal{N}(0, 1)$.
  }
  \label{fig:partition-gaussian}
\end{figure}

\subsection{Density Scoring Method}
\label{sec:densityscoring}
When talking about outliers---infrequent observations---the association with
probability can be quite intuitive. For instance, one would expect an anomaly
to happen rarely and be unlikely. Since the language of statistics often
associate the term {\em likelihood} with quantities like $p^{(\theta)}_X(x)$,
one might consider an unlikely sample to have a low "likelihood", that is, a low
probability density $p^*_X(x)$. Conversely, regular samples would have a high
density $p^*_X(x)$ following that reasoning. This is an intuition that is not
only prevalent in several modern anomaly detection
methods~\citep{bishop1994novelty, aabi2017panel, hendrycks2018deep,
kirichenko2020normalizing, rudolph2020same, liu2020energy} but also in techniques like
low-temperature sampling~\citep{graves2013generating} used for example in
\citet{kingma2018glow} and parmar et al. \cite{pmlr-v80-parmar18a}.

The associated approach,
described in~\citet{bishop1994novelty}, consists in 
defining the inliers as the points whose density exceed a certain threshold
$\lambda > 0$ (for example, chosen such that inliers include a predefined
amount of mass, e.g., $95\%$), making the modes the most regular points in
this setting.
$\Xout$ and $\Xin$ are then respectively the lower-level and upper-level sets
$\left\{x \in \mathcal{X}, p^*_X(x) \leq \lambda\right\}$ and
$\left\{x \in \mathcal{X}, p^*_X(x) > \lambda\right\}$
(see Figure \ref{fig:density-methods}b).
\end{paracol}
\nointerlineskip
\vspace{-6pt}
\begin{paracol}{2}
\switchcolumn

\subsection{Typicality Test Method}
\label{sec:typicality}
The {\em Gaussian Annulus theorem}~\citep{blum2016foundations}
generalized in \cite{vershynin2019high} attests that most of the mass of a
high-dimensional standard Gaussian $\mathcal{N}(0, \mathbb{I}_D)$ is located
close to the hypersphere of radius $\sqrt{D}$. However, the mode of its
density is at the center $0$. A natural conclusion is that the  {\em curse of
dimensionality} creates a discrepancy between the density upper-level sets and
what we expect as inliers 
\citep{choi2018waic, nalisnick2019detecting, morningstar2020density,
dieleman2020typicality}.
This motivated Nalisnick et al. \cite{nalisnick2019detecting} to propose another method for
testing whether a point is an inlier or not, relying on a measure of its {\em
typicality}.
This method relies on the notion of
{\em typical set}~\citep{cover1999elements}
defined by taking as inliers points whose average log-density is close to the
average log-density of the distribution (see Figure \ref{fig:density-methods}c).
\begin{Definition}[\cite{cover1999elements}]
Given independent and identically distributed elements
$\left(x^{(n)}\right)_{n\leq N}$ from a distribution with density $p^*_X$, the
typical set
$A_{\epsilon}^{(N)}(p^*_X) \subset \mathcal{X}^N$ is made of all sequences that satisfy
\begin{align*}
\abs{H(p^*_X) + \frac{1}{N} \sum_{n=1}^{N}{\log p^*_X\left(x^{(n)}\right)}} \leq
\epsilon,
\end{align*}
where $H(p^*_X)=-\mathbb{E}[\log p^*_X(X)]$ is the
(differential) entropy and $\epsilon > 0$ a constant. 
\end{Definition}
This method matches the intuition behind the Gaussian Annulus theorem on the
set of inliers of a high-dimensional standard Gaussian. Indeed, 
using a concentration inequality, we can show that
$\lim_{N \rightarrow +\infty}\left(P^*_{(X_i)_{1\leq n\leq
N}}\left(A_{\epsilon}^{(N)}\right)\right) = 1$, which means that with $N$ large
enough, $A_{\epsilon}^{(N)}(p^*_X)$ will contain most of
the mass of $(p^*_X)^N$, justifying the \mbox{name \textit{typicality}}.

\section{The Role of Reparametrization}
\label{sec:idea}
Density-based anomaly detection is applied in practice~\citep{bishop1994novelty, aabi2017panel, hendrycks2018deep,
kirichenko2020normalizing, rudolph2020same, liu2020energy}
as follows: first, learn a density estimator $p_X^{(\theta)}$ to approximate the data density $p_X^*$, and then plug that estimate in the density-based methods from Sections \ref{sec:densityscoring} and \ref{sec:typicality} to discriminate between inliers and outliers. Recent empirical failures  \citep{choi2018waic, nalisnick2018deep, hendrycks2018benchmarking} of this procedure applied to density scoring have been attributed to the discrepancy between $p_X^{(\theta)}$ and $p_X^*$ \citep{hendrycks2018deep,
Fetaya2020Understanding, morningstar2020density, kirichenko2020normalizing,
zhang2020hybrid}. Instead, we choose in this paper to question the fundamental assumption that these density-based methods should result in a correct classification between outliers and inliers. 
\subsection{A Binary Classification Analogy}
We start by studying the desired behavior of a classification method under infinite data and capacity, a setting where the user is provided with a perfect density model $p_X^{(\theta)} = p_X^*$.

\looseness=-1 In \citet{magritte1929trahison}, the author reminds us that the input $x$ we use is merely an arbitrary representation of the studied object (in other words, “a map is not the territory” \cite{korzybski1958science}), standardized here to enable the construction of a large-scale homogeneous dataset to train on \citep{hanna2020against}. This is after all the definition of a random variable $x = X(\omega)$, which is by definition a function from the underlying outcome $\omega$ to the corresponding observation $x$. For instance, in the case of object classification, $\omega$ is the object while $X(\omega)$ is the image (produced as a result of lighting, camera position and pose, lenses, and the image sensor). For images, a default representation is the bitmap one. However, this choice of representation remains arbitrary and practitioners have also trained their classifier using pretrained features instead~\citep{sermanet2013overfeat, simonyan2014very, szegedy2015going}, {\scshape Jpeg} representation~\citep{gueguen2018faster}, encrypted version of the data~\citep{xie2014crypto, gilad2016cryptonets}, or other resulting transformations $f(x) = f\big(X(\omega)\big)$, without modifying the associated labels. In particular, if $f$ is invertible, $f(x) = f\big(X(\omega)\big)$ contains the same information about $\omega$ as $x = X(\omega)$. Therefore both representations should be classified the same, as we associate the label with the underlying outcome $\omega$. If $c^*$ is the perfect classifier on $X$, then $c^* \circ f^{-1}$ should be the perfect classifier on $f(X) = (f \circ X)$ to assess the label of $\omega$, since $(c^* \circ f^{-1})\big(f(x)\big) = c^*(x)$.

As an illustration, we can consider the transition from of a cartesian coordinate system $(x_i)_{i\leq D}$ to a hyperspherical coordinate system, consisting of a radial coordinate $r>0$ and $(D-1)$ angular coordinates $(\phi_i)_{i < D}$,
\begin{align*}
\forall d<D,~x_d &= r\left(\prod_{i=1}^{d-1}{\sin(\phi_i)}\right)\cos(\phi_d) \\
x_D &= r\left(\prod_{i=1}^{D-2}{\sin(\phi_i)}\right)\sin(\phi_{D-1}),
\end{align*}
where for all $i\in \{1, 2, ..., D-2\},$ $\phi_i \in [0, \pi)$ and
$\phi_{D-1} \in [0, 2\pi)$. While significantly different, those two
systems of coordinates (or representations) describe the same vector and are connected by an invertible map $f$. In a Cartesian coordinate system, an optimal classifier $c^*(x) = \mathbbm{1}{(\sum x_i^2 > 1)}$ would become in a hyperspherical representation\linebreak $(c^* \circ f^{-1}) (r, \phi) = \mathbbm{1}{(r^2 > 1)}$.

The ability to learn the correct classification rule from infinite data and capacity, regardless of the representation used (or with minimal requirement), is a fundamental requirement (albeit weak) for a machine learning algorithm, and hence an interest in universal approximation properties, see \cite{cybenko1989approximation, hornik1991approximation, pinkus1999approximation}. While we do not dismiss the important role of the input representation as an inductive bias (e.g., using pretrained features as inputs), its influence should in principle dissipate entirely in the infinite data and capacity regime and the resulting solution from this ideal setting should be unaffected by this inductive bias. In ideal conditions, solutions to classification should be invariant to any invertible change \mbox{of representation}.

We consider this is in fact one of key tenets behind deep learning~\citep{goodfellow2016deep} and feature engineering/learning in general \citep{bengio2013representation}.

\subsection{A Principle for Anomaly Detection Methods}

Current practices of deep anomaly detection commonly include the use of deep density models on either default input feature~\citep{choi2018waic, nalisnick2018deep, hendrycks2018deep,
kirichenko2020normalizing, rudolph2020same, nalisnick2019detecting}
or features learned independently from the anomaly detection task~\citep{lee2018simple, krusinga2019understanding, morningstar2020density,
winkens2020contrastive}.
The process of picking a particular input representation is rarely justified in the context of density-based anomaly detection, which suggests that a similar implicit assumption is being used: \textit{the status of inlier/outlier corresponds to the underlying outcome $\omega$ behind an input feature $x = X(\omega)$, whose only role is to inform us on $\omega$}.
 \looseness=-1 As described in~Section \ref{sec:pb}, the goal of anomaly detection is, like classification, to discriminate (although generally in an unsupervised way) between inliers and outliers.
Similarly to classification, the label of inlier/outlier of an underlying outcome should remain invariant to reparametrization in an infinite data and capacity setting, especially since information about $\omega$ (and whether the outcome is anomalous or not) is  conserved under an invertible transformation up to numerical instabilities, see \cite{behrmann2020understanding}. We consider the following principle: 
\begin{principle}
In an infinite data and capacity setting, the result of an anomaly detection
method should be invariant to any continuous invertible reparametrization $f$.
\end{principle}
This principle is coherent with the fact that, with $f$ invertible, 
the set of outliers $\Xout$ remains a low probability subset as 
$P_X(\Xout) = P_{f(X)}\big(f(\Xout)\big)$ and
$\forall x \in \mathcal{X},~x \in \Xout \iff f(x) \in f(\Xout)$.
 \looseness=-1 However, density-based methods do not follow this principle as densities are not representation-invariant.
In particular, the change of variables formula \citep{kaplan1952advanced}, also used in Dinh et al. \cite{tabak2013family}, Tabak and Turner \cite{dinh2014nice}, Rezende and Mohamend \cite{pmlr-v37-rezende15},
formalizes a simple intuition of this behavior: where points are brought closer
together the density increases whereas this density decreases when points are
spread apart.

The formula itself is written as:
\[p^*_{f(X)}\big(f(x)\big) = p^*_{X}(x)\abs{\frac{\partial f}{\partial x^T}(x)}^{-1}\]
where $\abs{\frac{\partial f}{\partial x^T}(x)}$ is the Jacobian determinant
of $f$ at $x$, a quantity that reflects a local change in volume incurred by
$f$.
Figure \ref{fig:change-of-variables} already illustrates how the function $f$
Figure \ref{fig:change-of-variables}b can spread apart points close to the
extremities to decrease the corresponding density around $x=0$ and $x=1$,
and, as a result, turns the density on the left Figure \ref{fig:change-of-variables}a into the
density on the right Figure \ref{fig:change-of-variables}c.
Figure \ref{fig:change-of-coord} shows how much a simple change of coordinate
system, from Cartesian (Figure \ref{fig:change-of-coord}a) to hyperspherical
(Figure \ref{fig:change-of-coord}b), can significantly affect the resulting density
associated with a point. This comes from the Jacobian
determinant of this change of coordinates:
\[r^{D-1}\left(\prod_{d=1}^{D-1}{\big(\sin(\phi_d)\big)^{D - d - 1}}\right).\]

With these examples, one can wonder
to which degree an invertible change of representation can affect the density
and thus the anomaly detection methods presented in Sections 
\ref{sec:densityscoring} and \ref{sec:typicality} that use it. This is what we explore in Section \ref{sec:chofvar}.
\vspace{-6pt}
\begin{figure}[H]

    \begin{subfigure}[t]{0.47\textwidth}
\includegraphics[scale=1]{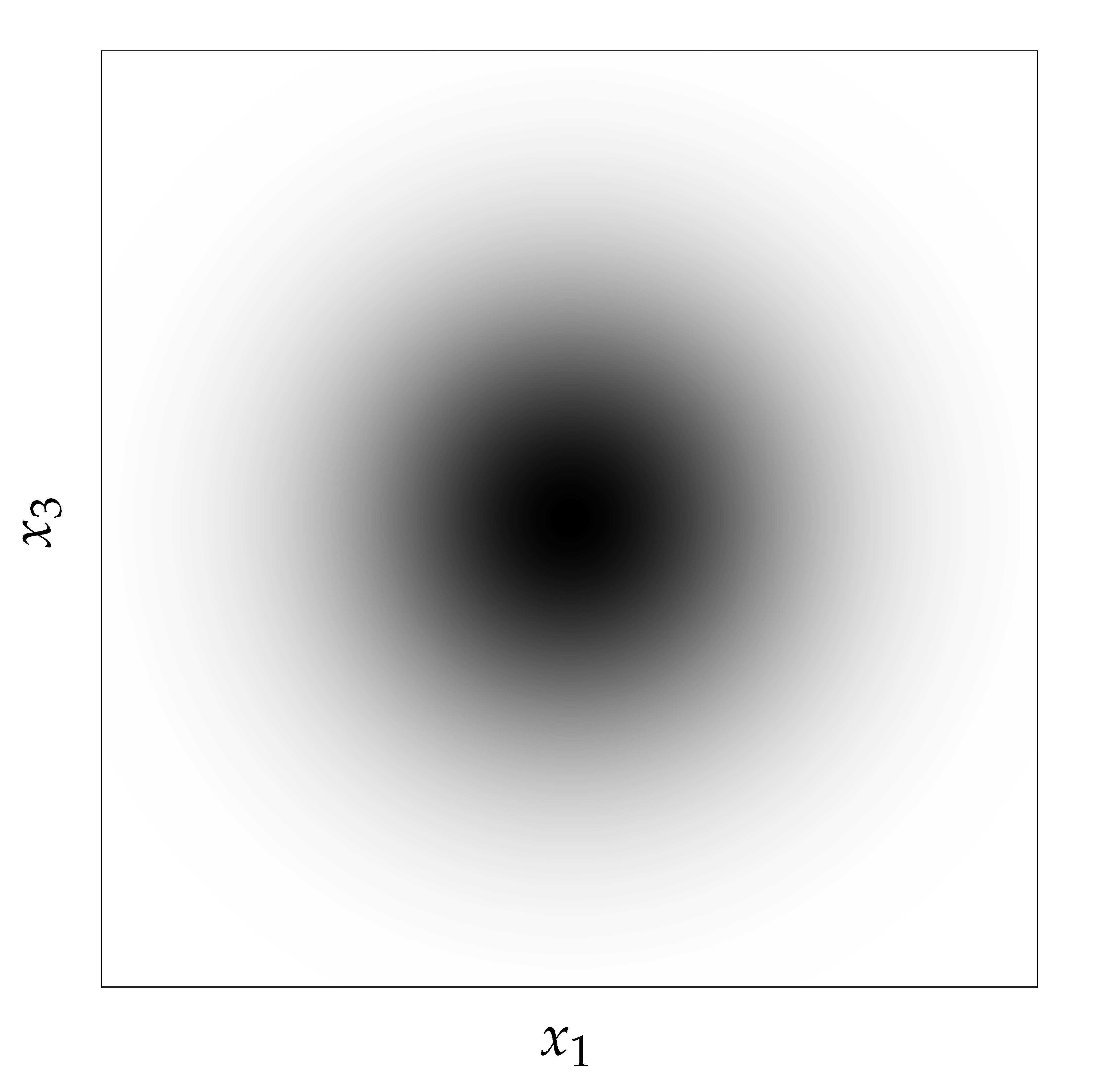}
  \caption{}
  \label{fig:cartesian_pdf}
  \end{subfigure}
  \hspace{-45pt}
  \begin{subfigure}[t]{0.47\textwidth}
 \includegraphics[scale=1]{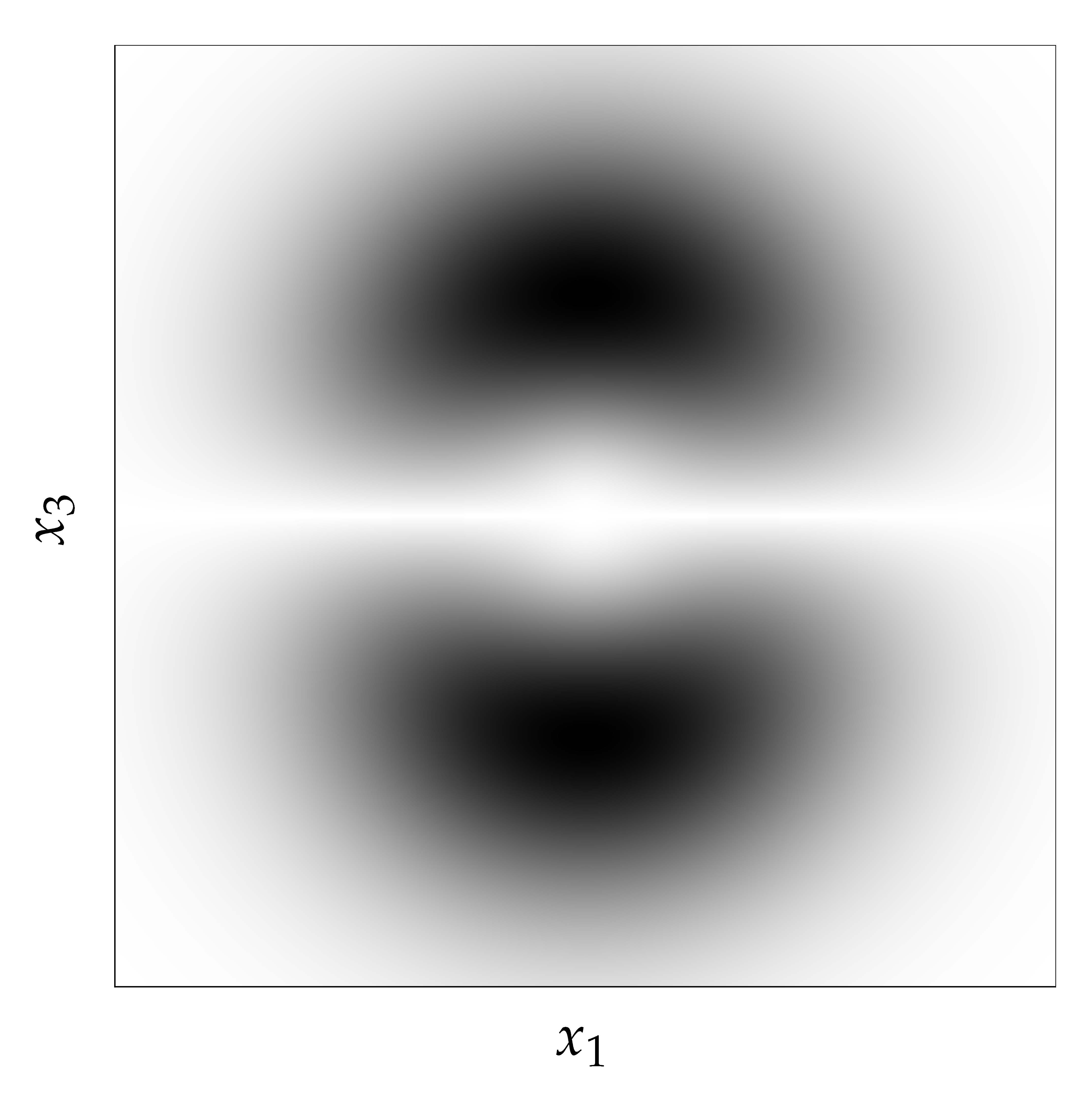}
  \caption{}
  \label{fig:polar_pdf}
  \end{subfigure}
  \vspace{9pt}
  \caption{Illustration of the change of variables formula for a three-dimensional standard Gaussian distribution with a change of coordinate system, from Cartesian to hyperspherical
  (where density follows the intuition of the Gaussian Annulus Theorem) (\textbf{a}) A three-dimensional standard Gaussian distribution density in Cartesian coordinates on the hyperplane defined by $x_2 = 0$. (\textbf{b}) A three-dimensional standard Gaussian distribution density in hyperspherical coordinates (plotted in Cartesian coordinates) on the hyperplane defined by $x_2 = 0$.}
  \label{fig:change-of-coord}
\end{figure}

\section{Leveraging the Change of Variables Formula}
\label{sec:chofvar}
\subsection{Uniformization}
\label{sec:uniformization}

We start by showing that unambiguously defining outliers and inliers with any
density-based approach becomes impossible when considering a particular type
of invertible reparametrization of the problem, irrespective of dimensionality.

Under weak assumptions, one can map any distribution to a uniform distribution 
using an invertible transformation~\citep{hyvarinen1999nonlinear}.
This is in fact a common strategy for sampling from complicated one-dimensional distributions
\citep{devroye1986sample}. Figure \ref{fig:uniformization} shows an example of
this where a bimodal distribution (Figure \ref{fig:uniformization}a) is pushed through
an invertible map (Figure \ref{fig:uniformization}b) to obtain a uniform distribution
(Figure \ref{fig:uniformization}c). 

%\clearpage
\end{paracol}
\nointerlineskip
\begin{figure}[H]
\widefigure
  \centering
  \begin{subfigure}[t]{0.3\textwidth}
  \centering
  \begin{tikzpicture}
    \node (img)  {\includegraphics[width=.9\textwidth]{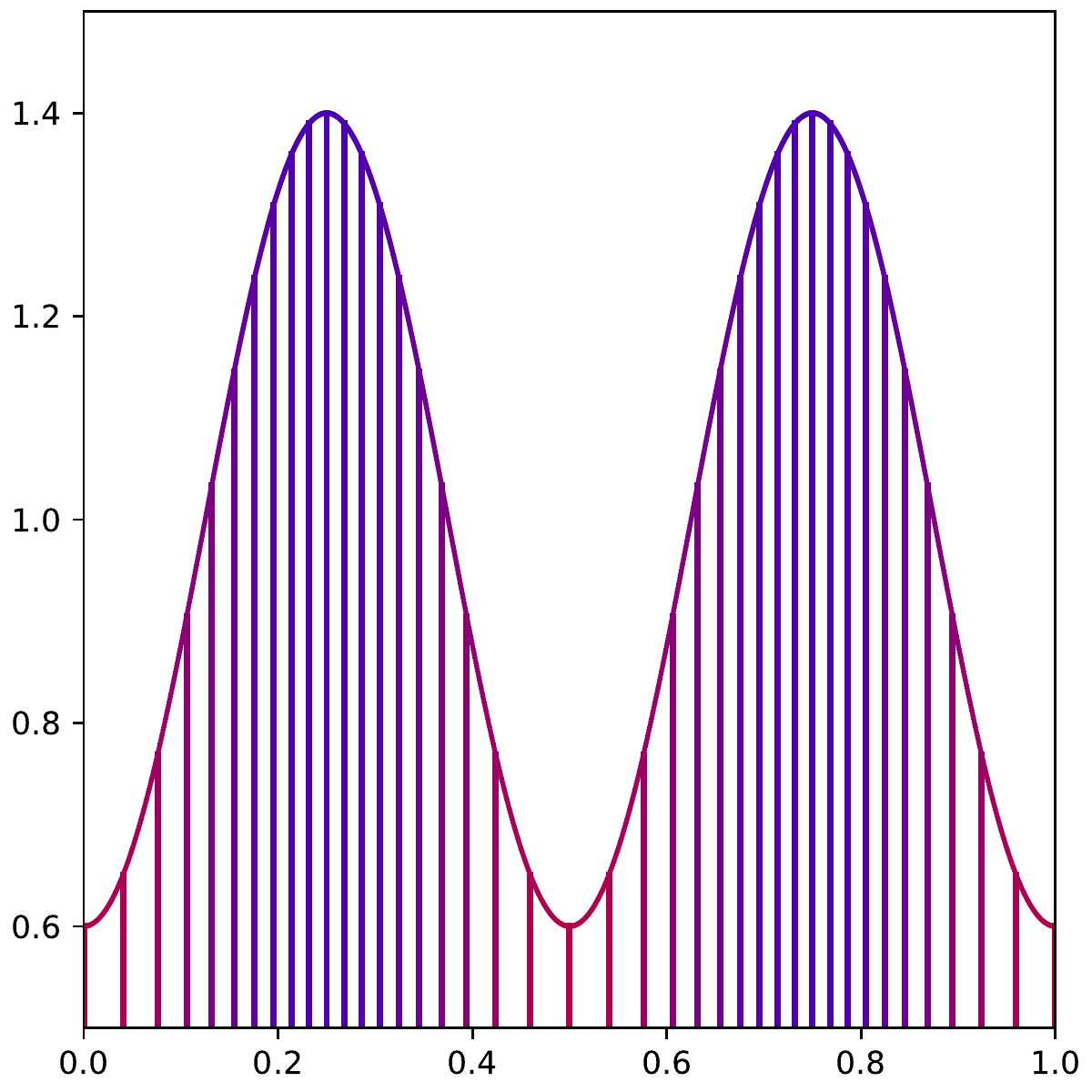}};
    \node[below of=img, node distance=2.9cm, yshift=0cm] {\small $x$};
    \node[left of=img, node distance=2.9cm, rotate=90, anchor=center] {\small $p^*_X(x)$};
  \end{tikzpicture}
  \caption{}
  \label{fig:pdf_2_unif}
  \end{subfigure}
  \begin{subfigure}[t]{0.3\textwidth}
  \centering
  \begin{tikzpicture}
    \node (img)  {\includegraphics[width=.9\textwidth]{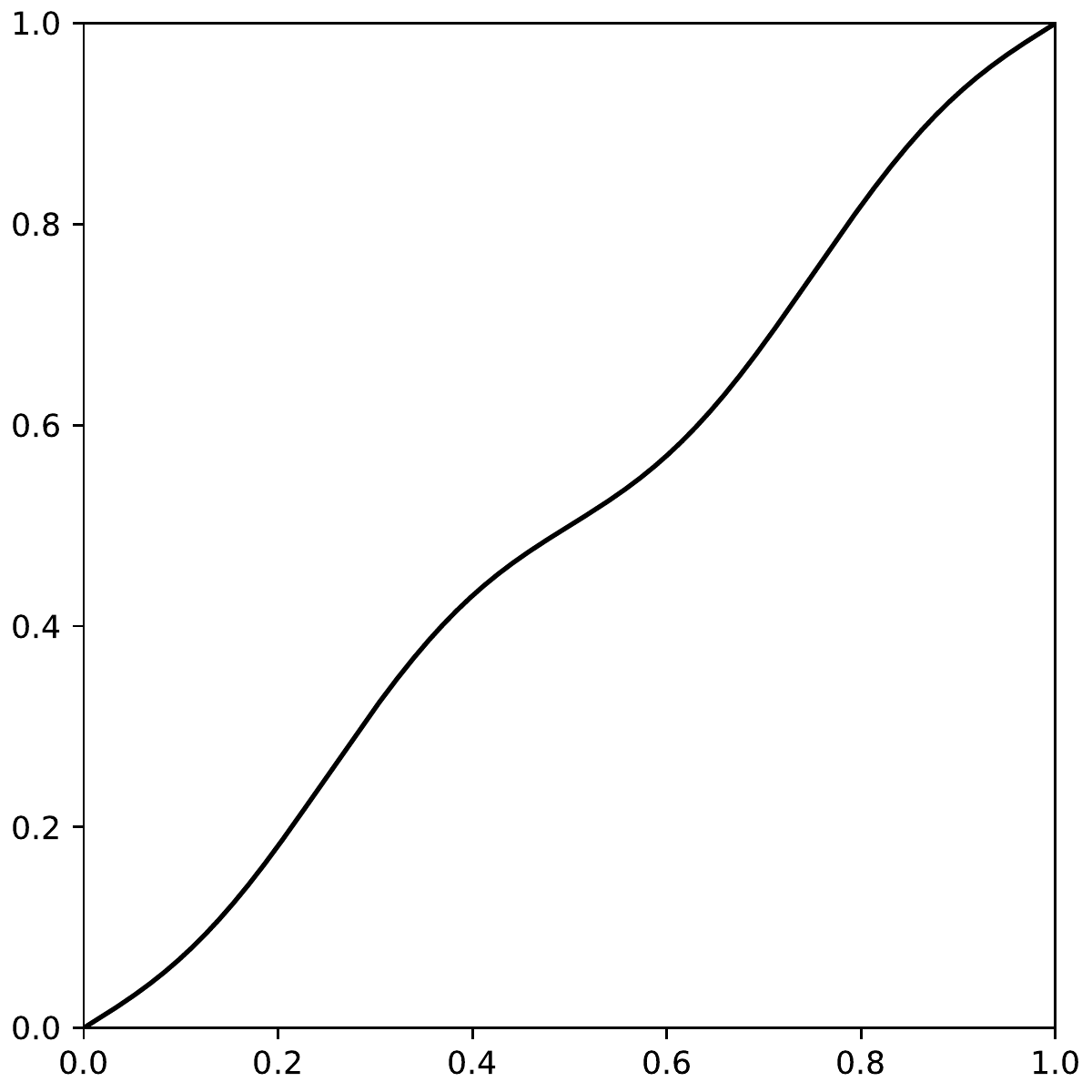}};
    \node[below of=img, node distance=2.9cm, yshift=0cm] {\small $x$};
    \node[left of=img, node distance=2.9cm, rotate=90, anchor=center] {\small $CDF_{p^*_X}(x)$};
  \end{tikzpicture}
  \caption{}
  \label{fig:cdf_2_unif}
  \end{subfigure}
  \begin{subfigure}[t]{0.3\textwidth}
  \centering
  \begin{tikzpicture}
    \node (img)  {\includegraphics[width=.9\textwidth]{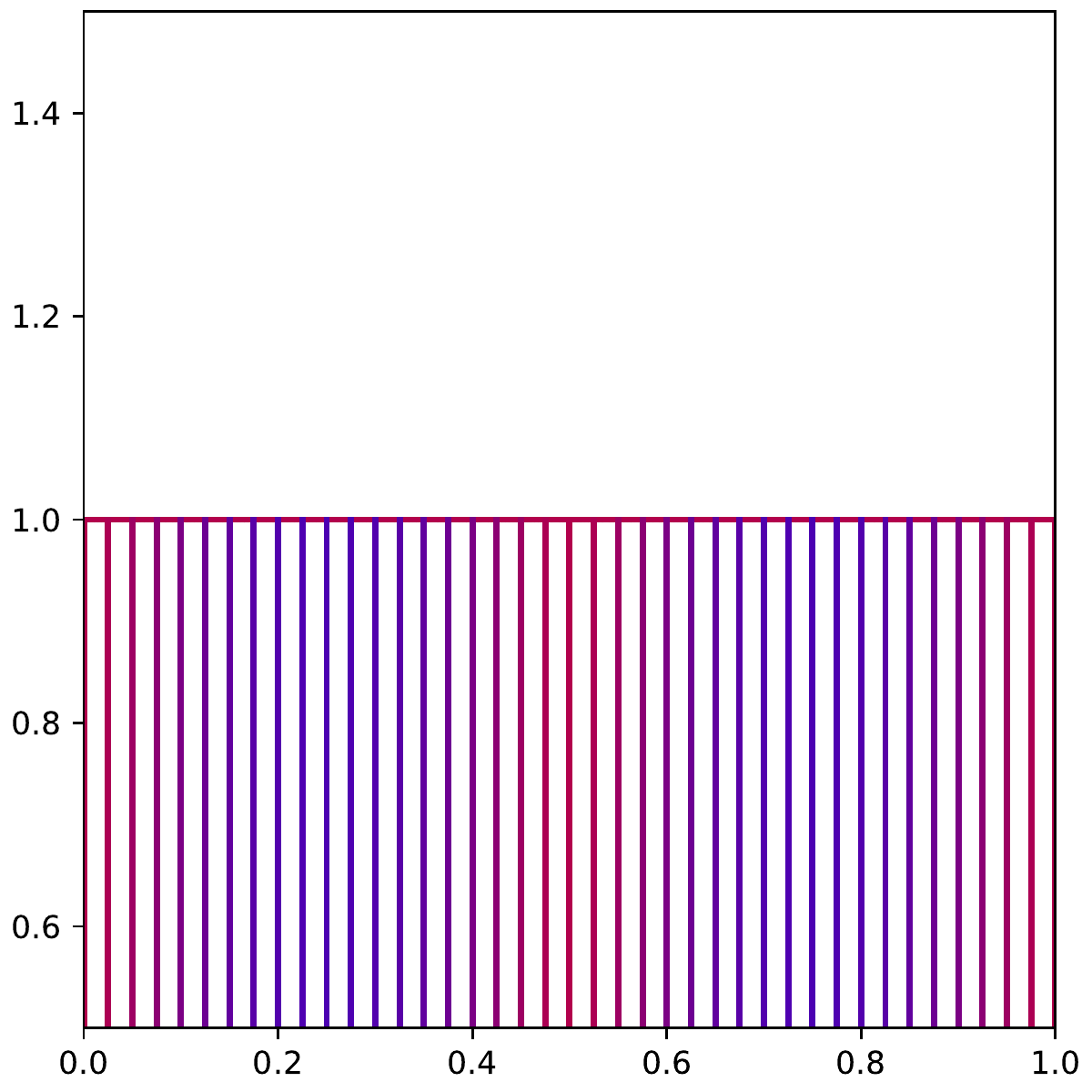}};
    \node[below of=img, node distance=2.9cm, yshift=0cm] {\small $CDF_{p^*_X}(x)$};
    \node[left of=img, node distance=2.9cm, rotate=90, anchor=center] {\small $p^*_{CDF_{p^*_X}(X)}\big(CDF_{p^*_X}(x)\big)$};
  \end{tikzpicture}
  \caption{}
  \label{fig:uniform_pdf}
  \end{subfigure}
  \vspace{6pt}
  \caption{Illustration of the one-dimensional case version
  of a Knothe-Rosenblatt rearrangement, which is just the application of
  the cumulative distribution function $CDF_{p^*_X}$ on the variable $x$.
  Points $x$  with high density $p^*_X(x)$ are in blue and points with low density
  $p^*_X(x)$ are in red. (\textbf{a}) An example of a distribution density $p^*_X$. (\textbf{b}) The corresponding cumulative distribution function $CDF_{p^*_X}$. (\textbf{c}) The resulting density from applying $CDF_{p^*_X}$ to $X \sim p^*_X$
  is $p^*_{CDF_{p^*_X}(X)} = \mathcal{U}([0, 1])$.}
  \label{fig:uniformization}
\end{figure}
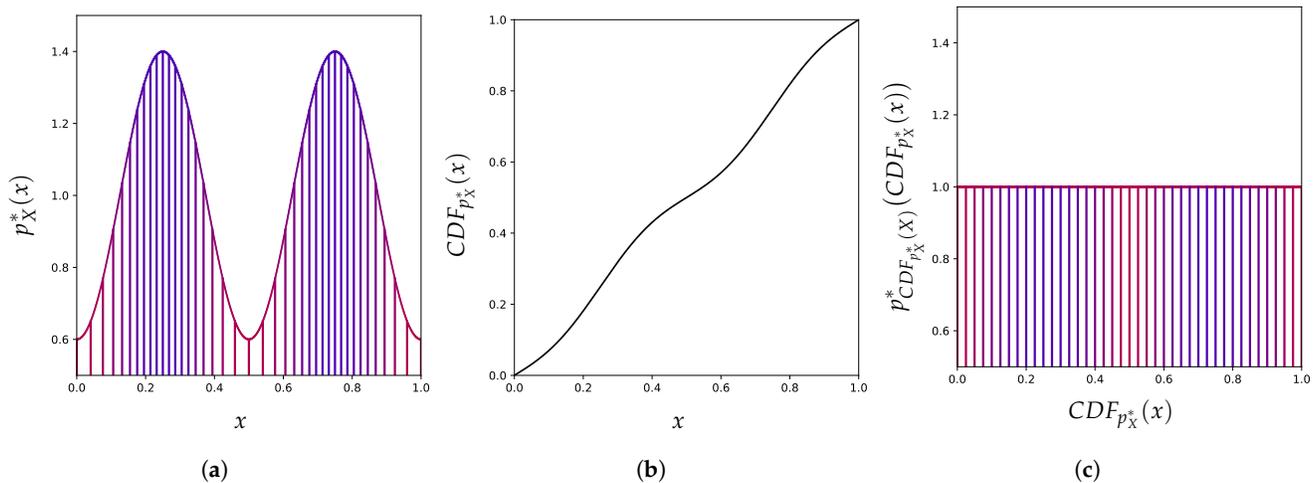
\begin{paracol}{2}
\switchcolumn

To construct this invertible uniformization function,
we rely on the notion of Knothe-Rosenblatt rearrangement
\citep{rosenblatt1952remarks, knothe1957contributions}.
A Knothe-Rosenblatt rearrangement notably used \mbox{in 
\cite{hyvarinen1999nonlinear}} is defined for
a random variable $X$ distributed according to a strictly positive
density $p^*_X$ with a convex support $\mathcal{X}$,
as a continuous invertible map
$f^{(KR)}$ from $\mathcal{X}$ onto $[0, 1]^D$ such that
$f^{(KR)}(X)$ follows a uniform distribution in this hypercube.
This rearrangement is constructed as follows:
$\forall d \in \{1, ..., D\}, f^{(KR)}(x) = CDF_{p^*_{X_d \mid X_{<d}}}(x_d \mid x_{<d})$
where $CDF_{p}$ is the cumulative distribution function corresponding to the
density $p$.

In these new coordinates, neither the density scoring method nor the
typicality test approach can discriminate between inliers and outliers in this
uniform $D$-dimensional hypercube $[0, 1]^D$. 
Since the resulting density $p^*_{f^{(KR)}(X)} = 1$
is constant, the density scoring method attributes the same regularity to every
point or set of points. Moreover, a typicality test on $f^{(KR)}(X)$ will always succeed as
\begin{align*}
\forall \epsilon > 0, N \in \mathbb{N}^*, \forall \left(x^{(n)}\right)_{n \leq N},
T_{typ}\left(p_{f^{(KR)}(X)}^*, \big(x^{(n)}\big)_{n \leq N}\right)
= 0 \leq \epsilon.
\end{align*}

However, these uniformly distributed points are merely a
different representation of the same initial points. Therefore, if the identity of the
outliers is ambiguous in this uniform distribution, then anomaly detection in
general should be as difficult.

\subsection{Arbitrary Scoring}
\label{sec:arbitraryscoring}
We find that it is
possible to build a reparametrization of the problem to impose to each point an
arbitrary density level in the new representation. To illustrate this idea,
consider some points from a distribution whose density is depicted in
Figure \ref{fig:arbitrary-scoring}a and a score function indicated in red in
Figure \ref{fig:arbitrary-scoring}b. In this example, high-density regions
correspond to areas with low score value (and vice-versa), such that the ranking from the densities is reversed with this new score. Given that desired score function, we show how to systematically build
a reparametrization (depicted in Figure \ref{fig:arbitrary-scoring}c) such that the
density in this new representation (Figure \ref{fig:arbitrary-scoring}d) now matches the
desired score, which can be designed to mislead density-based methods
into a wrong classification of anomalies by modifying a single dimension (in a potentially high-dimensional input vector).

%\clearpage
\end{paracol}
\nointerlineskip
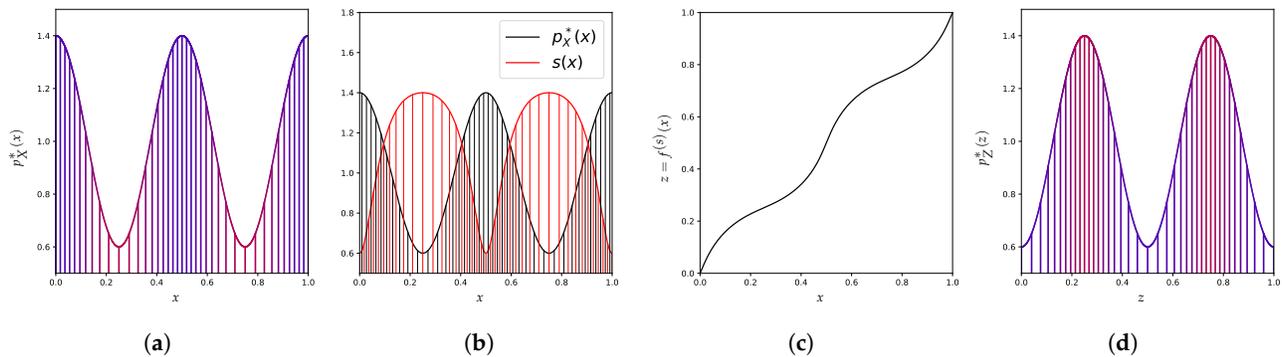
\begin{figure}[H]
\widefigure
  \centering
  \begin{subfigure}[t]{0.22\textwidth}
  \centering
  \begin{tikzpicture}
    \node (img)  {\includegraphics[width=.9\textwidth]{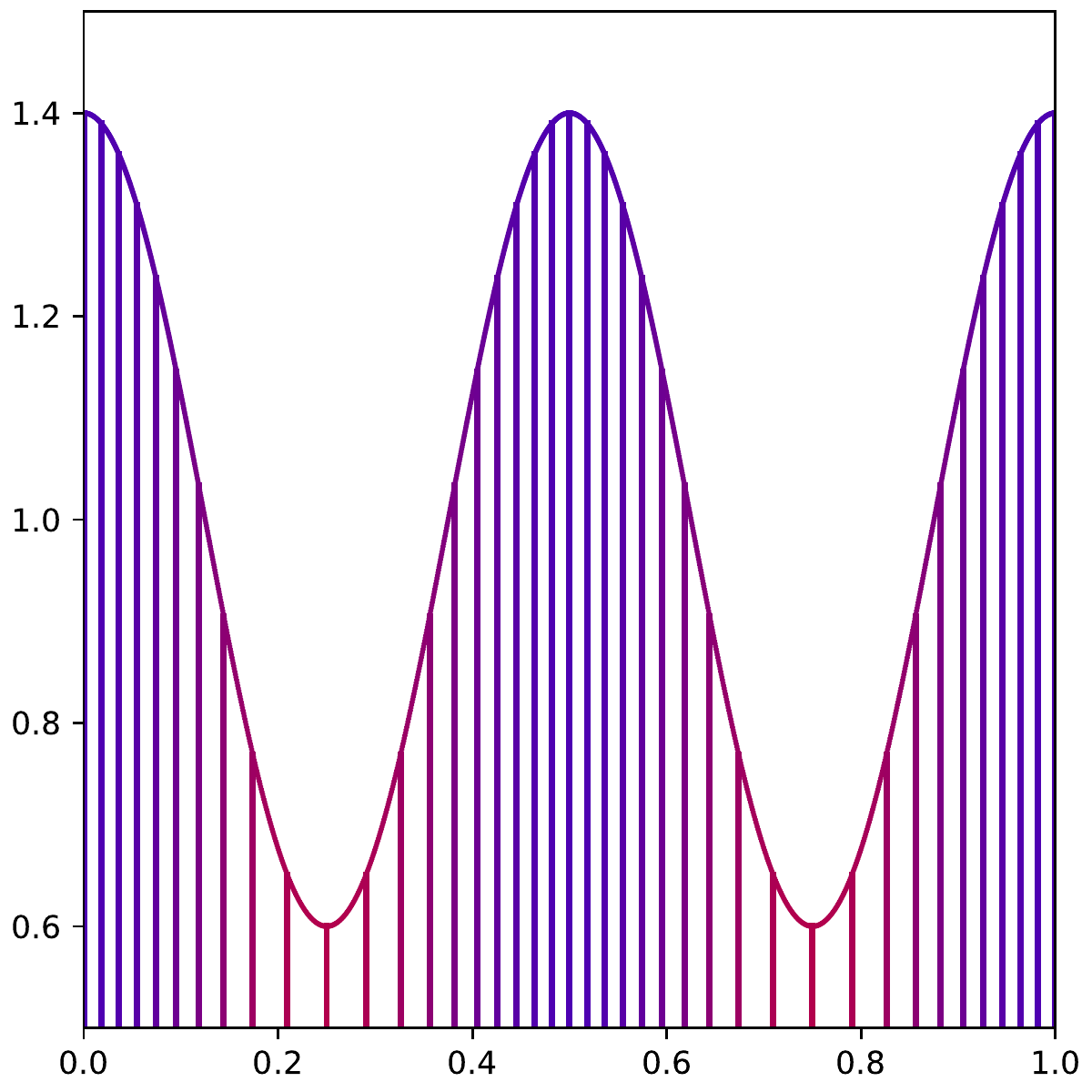}};
    \node[below of=img, node distance=2.cm, yshift=0cm] {\tiny $x$};
    \node[left of=img, node distance=2.1cm, rotate=90, anchor=center] {\tiny $p^*_X(x)$};
  \end{tikzpicture}
  \caption{}
  \label{fig:pdf_1_as}
  \end{subfigure}
  \begin{subfigure}[t]{0.22\textwidth}
  \centering
  \begin{tikzpicture}
    \node (img)  {\includegraphics[width=.9\textwidth]{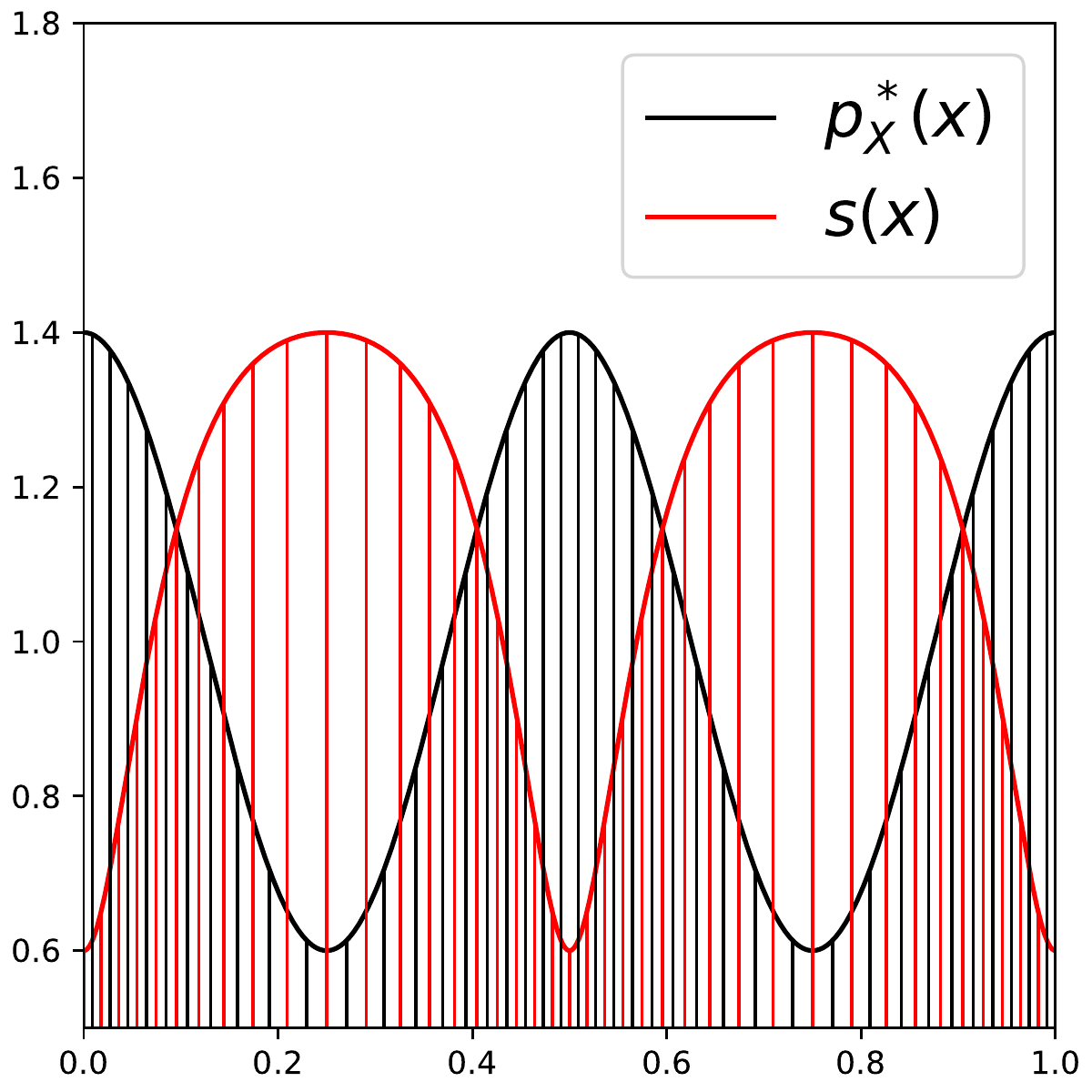}};
    \node[below of=img, node distance=2.cm, yshift=0cm] {\tiny $x$};
    \node[left of=img, node distance=1.6cm, rotate=90, anchor=center] {\tiny $~$};
  \end{tikzpicture}
  \caption{}
  \label{fig:pdf_sx_as}
  \end{subfigure}
  \begin{subfigure}[t]{0.22\textwidth}
  \centering
  \begin{tikzpicture}
    \node (img)  {\includegraphics[width=.9\textwidth]{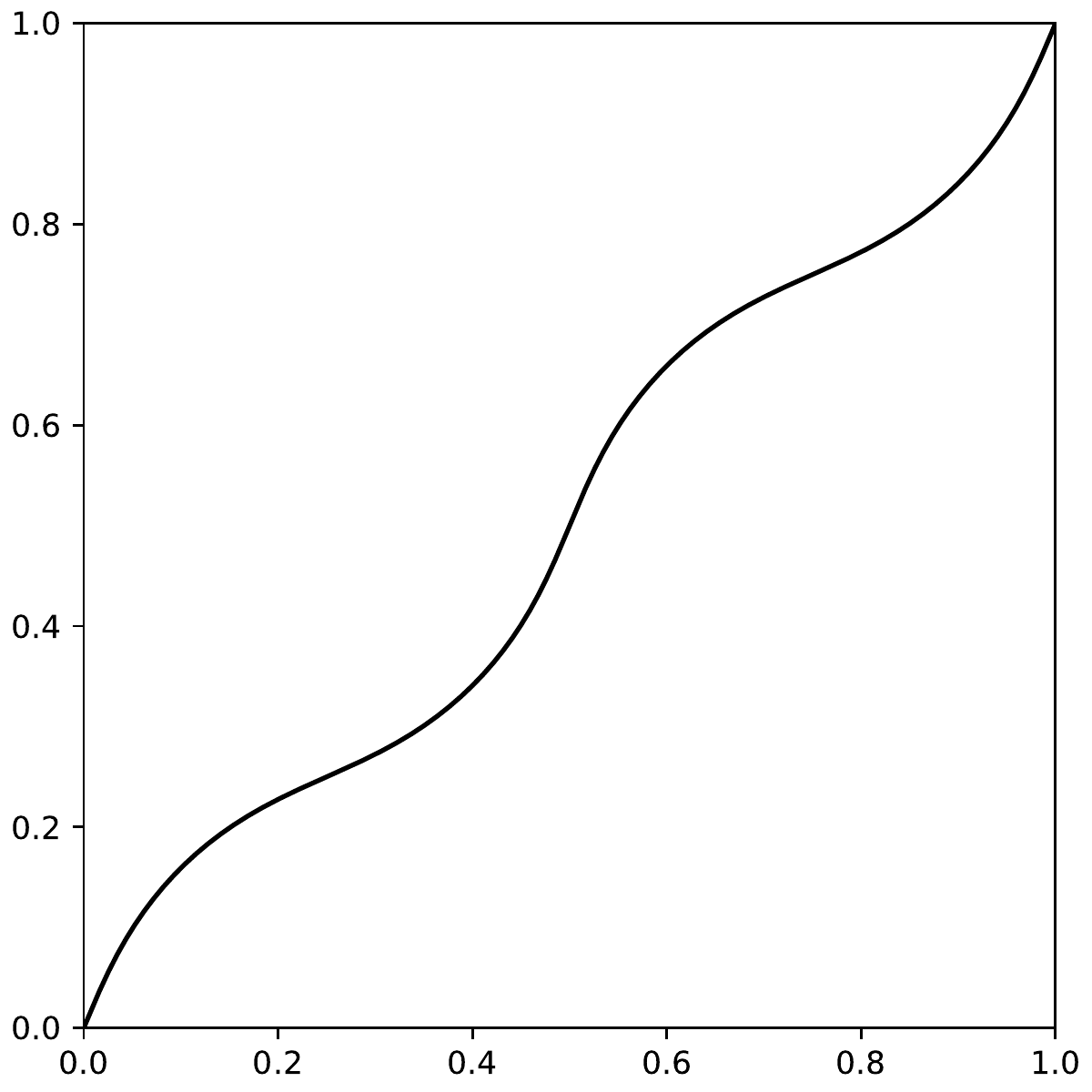}};
    \node[below of=img, node distance=2.cm, yshift=0cm] {\tiny $x$};
    \node[left of=img, node distance=2.1cm, rotate=90, anchor=center] {\tiny $z = f^{(s)}(x)$};
  \end{tikzpicture}
  \caption{}
  \label{fig:pdf_1_to_pdf_2_as}
  \end{subfigure}
  \begin{subfigure}[t]{0.22\textwidth}
  \centering
  \begin{tikzpicture}
    \node (img)  {\includegraphics[width=.9\textwidth]{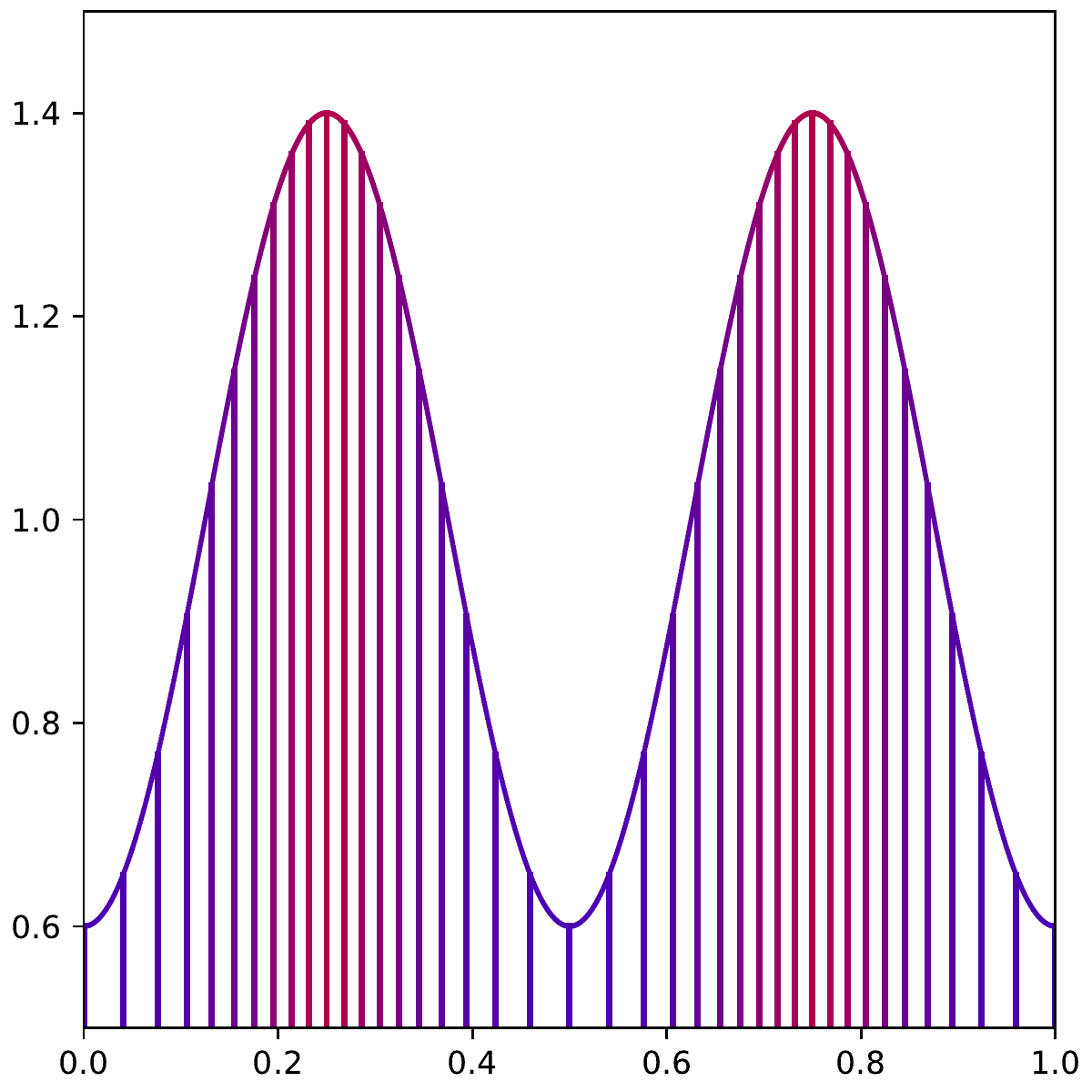}};
    \node[below of=img, node distance=2.cm, yshift=0cm] {\tiny $z$};
    \node[left of=img, node distance=2.1cm, rotate=90, anchor=center] {\tiny $p^*_{Z}(z)$};
  \end{tikzpicture}
  \caption{}
  \label{fig:pdf_2_as}
  \end{subfigure}
  \vspace{6pt}
  \caption{Illustration of how we can modify the space with an invertible
  function so that each point $x$ follows a predefined score. In
  (\textbf{a},\textbf{b}) points with high original density $p^*_X(x)$ are in blue and red for low original density. (\textbf{a}) An example of a distribution density $p^*_X$. (\textbf{b}) The density $p^*_X$ (in black) and the desired density
  scoring $s$ (in red). (\textbf{c}) A continuous invertible reparametrization $z = f^{(s)}(x)$ such that
  $p^*_{Z}(z) = s(x)$. (\textbf{d}) Resulting density $p^*_{Z}$ from applying $f^{(s)}$ to
  $X \sim p^*_X$ as a function of $z = f^{(s)}(x)$.}
  \label{fig:arbitrary-scoring}
\end{figure}
\vspace{-15pt}
\begin{paracol}{2}
\switchcolumn

\begin{restatable}{Proposition}{propone}
\label{proposition:arbitraryscoring}
For any random variable $X \sim p^*_X$ with $p^*_X$  strictly positive (with $\mathcal{X}$ convex) and any
measurable continuous function $s: \mathcal{X} \rightarrow \mathbb{R}_{+}^*$ bounded
below by a strictly positive number, there exists a continuous bijection
$f^{(s)}$ such that for any $x \in \mathcal{X},
p^*_{f^{(s)}(X)}\big(f^{(s)}(x)\big) = s(x)$.
\end{restatable}
\begin{proof}
\vspace{-1.em}
We write $x$ to denote $(x_1, \dots, x_{D-1}, x_D)$ and $(x_{<D}, t)$
for $(x_1, \dots, x_{D-1}, t)$. Let
$f^{(s)}:\mathcal{X} \rightarrow \mathcal{Z} \subset \R^D$ be a function such
that \[\big(f^{(s)}(x)\big)_D = \int_{0}^{x_D}{\frac{p^*_X\big((x_{<D}, t)\big)}{s\big((x_{<D}, t)\big)}dt},\]
and $\forall d \in \{1, ..., D-1\}, \big(f^{(s)}(x)\big)_d = x_d$.
As $s$ is bounded below, $f^{(s)}$ is well defined and invertible.
\end{proof}

By the change of variables formula,
\begin{align*}
\forall x \in \mathcal{X},~p^*_{f^{(s)}(X)}\big(f^{(s)}(x)\big) =
p^*_X(x)
\cdot \abs{\frac{\partial f^{(s)}}{\partial x^T}(x)}^{-1}=p^*_X(x)
\cdot \left(\frac{p^*_X(x)}{s(x)}\right)^{-1} = s(x).
\end{align*}

If $\Xin$ and $\Xout$ are respectively the true sets of inliers and outliers,
we can pick a ball $A \subset \Xin$ such that $P^*_X(A) = \alpha < 0.5$, we can
choose $s$ such that for any $x \in (\mathcal{X} \setminus A)$, \linebreak$s(x) = 1$ and
for any $x \in A, s(x) = 0.1$. With this choice of $s$
(or a smooth approximation) and the function
$f^{(s)}$
defined earlier, both the density scoring and the (one-sample) typical set
methods will consider the set of inliers to be $(\mathcal{X} \setminus A)$
while $\Xout \subset (\mathcal{X} \setminus A)$, making their results
completely wrong.
While we can also reparametrize the problem so that these methods may succeed, e.g., a parametrization where anomalies have low density for the density scoring method,
such a reparametrization \textit{requires knowledge of $(p^*_X / s)(x)$}.
Without any constraint on the space considered, individual densities can be
arbitrarily manipulated, which reveals how little this quantity says about
the underlying outcome in general.

\subsection{Canonical Distribution}
\label{sec:invariant}

Our analysis from Section \ref{sec:arbitraryscoring} revealing
that densities or low typicality regions are not sufficient conditions for an
observation to be an anomaly, whatever its distribution or its dimension, we
are now interested in investigating whether additional stronger assumptions
can lead to some guarantees for anomaly detection. Motivated by several
representation learning algorithms which attempt to learn a mapping to a
predefined distribution, e.g., a standard Gaussian, see \cite{
NIPS2000_1856, kingma2013auto, rezende2014stochastic, dinh2014nice,
krusinga2019understanding}, we
consider the more restricted setting of a fixed distribution of our choice,
whose regular regions could for instance be known. Surprisingly, we find that
it is possible to exchange the densities of an inlier and an outlier even
within a canonical distribution.

\begin{restatable}{Proposition}{proptwo}
\label{prop:invariant}
For any strictly positive density function $p^*_X$ over a
convex space $\mathcal{X}\subseteq\mathbb{R}^D$ with $D\geq2$, for any
$x_{in}, x_{out}$ in the interior $\mathcal{X}^{\mathrm{o}}$ of $\mathcal{X}$,
there exists a continuous bijection $f:\mathcal{X} \rightarrow \mathcal{X}$
such that $p^*_X = p^*_{f(X)}$,
$p^*_{f(X)}\left(f\left(x^{(in)}\right)\right) = p^*_X\left(x^{(out)}\right)$,
and
$p^*_{f(X)}\left(f\left(x^{(out)}\right)\right) = p^*_X\left(x^{(in)}\right)$.
\end{restatable}

\begin{proof}
The proof is given in Appendix \ref{app:proof}. It relies on the transformation depicted in
\mbox{Figure \ref{fig:f-rotation}}, which can swap two points while acting in a very local
area. If the distribution of points is uniform inside this local area, then
this distribution will be unaffected by this transformation. To come to this, we use the uniformization method presented in~\citep{hyvarinen1999nonlinear}, along with a linear function to fit this local area
inside the support of the distribution (see \mbox{Figure \ref{fig:put-ball-in}}). Once
those two points have been swapped, we can reverse the functions preceding
this swap to recover the original distribution overall.
\end{proof}

\begin{figure}[H]
  \begin{subfigure}[t]{0.48\textwidth}
  \includegraphics[width=.5\textwidth]{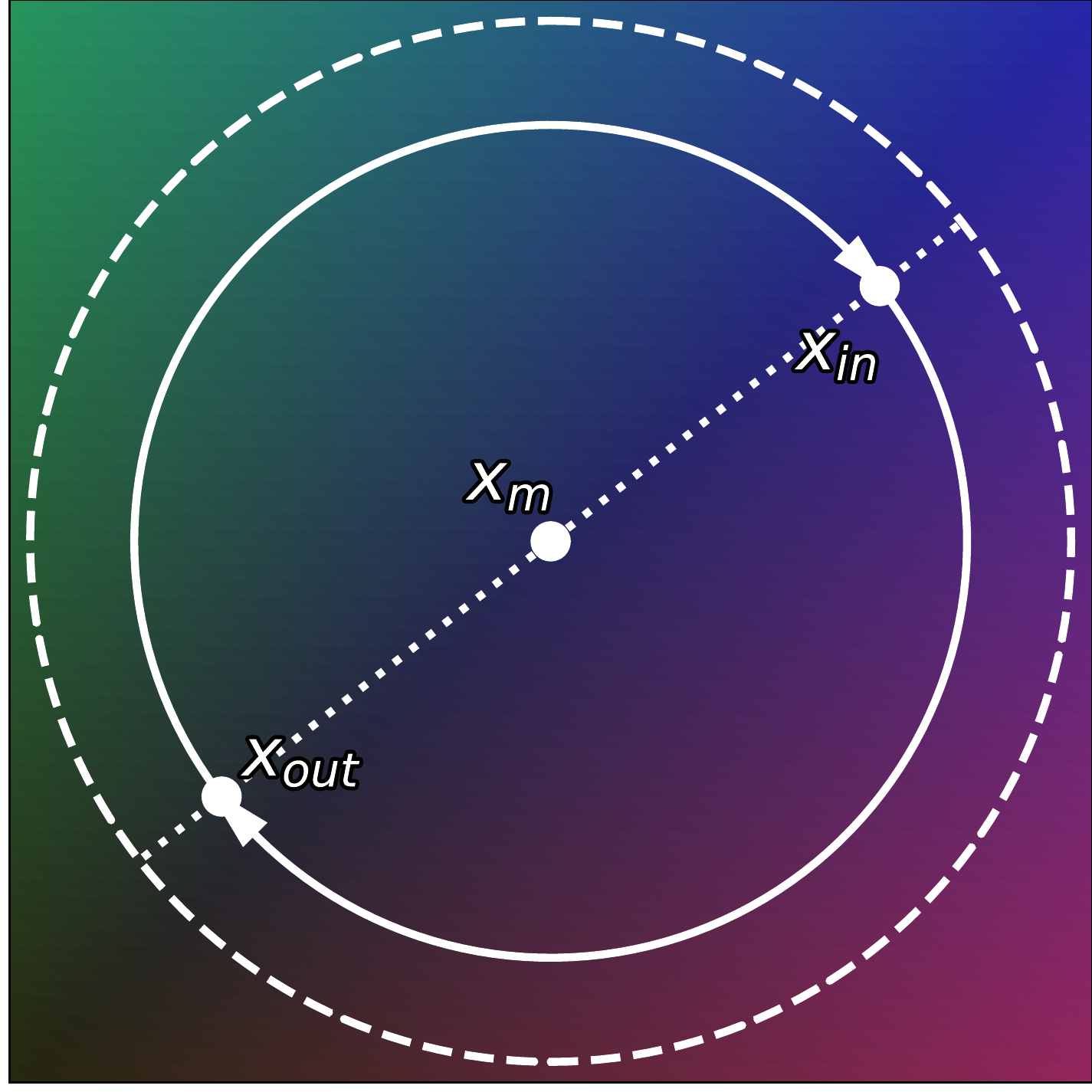}
  \caption{}
  \label{fig:f-rot-0}
  \end{subfigure}
  \hspace{-30pt}
  \begin{subfigure}[t]{0.48\textwidth}
  \includegraphics[width=.5\textwidth]{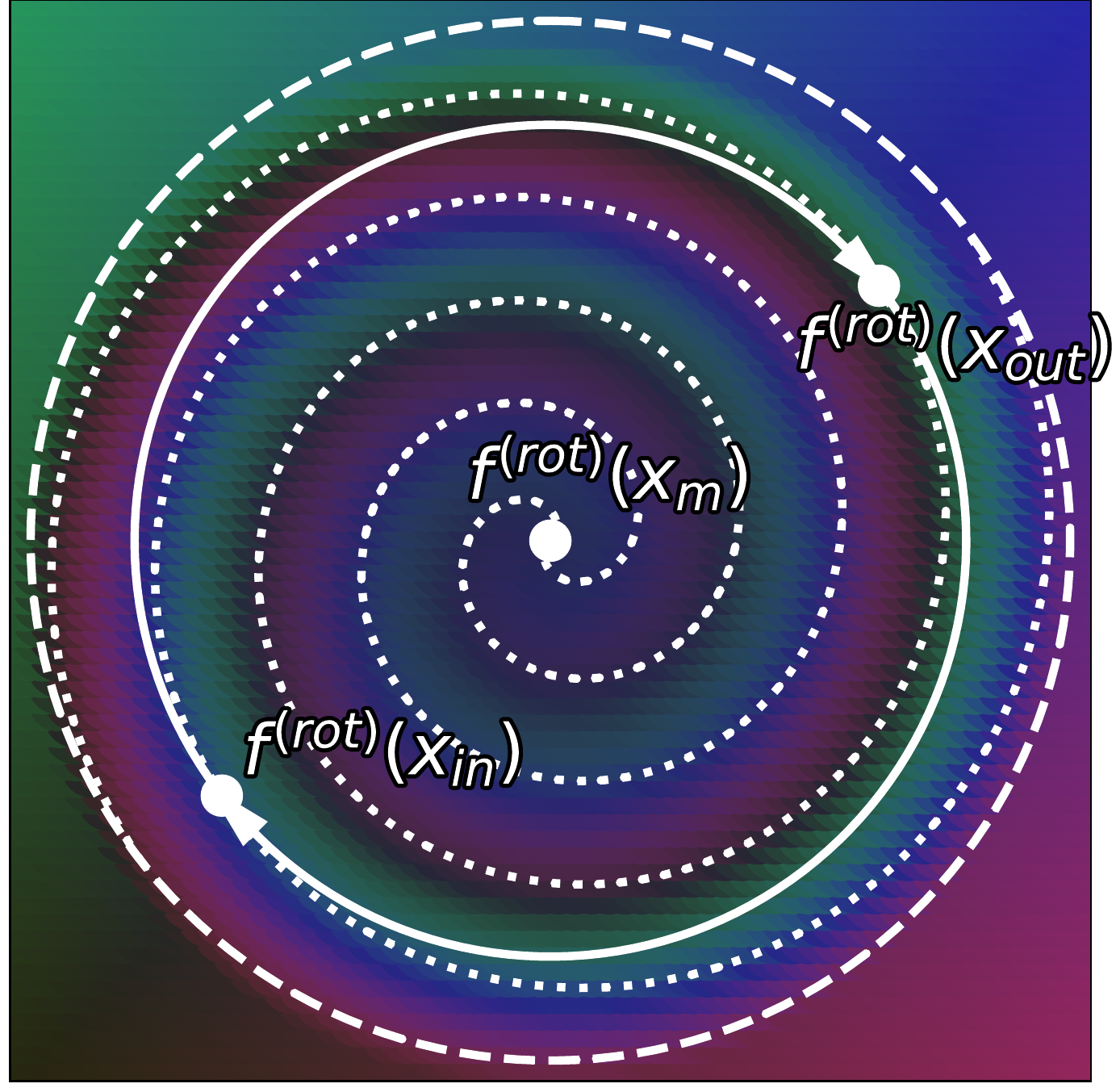}
  \caption{}
  \label{fig:f-rot-1}
  \end{subfigure} 
  \vspace{6pt}
  \caption{Illustration of the norm-dependent rotation, a locally-acting
  bijection that allows us to swap two different points while preserving a
  uniform distribution (as a volume-preserving function).\linebreak (\textbf{a}) Points $x_{in}$ and $x_{out}$ in a uniformly distributed subset.
  $f^{(rot)}$ will pick a two-dimensional plane and use the polar coordinate
  using the mean $x_m$ of $x_{in}$ and $x_{out}$ as the center. (\textbf{b}) Applying a bijection $f^{(rot)}$ exchanging the points $x_{in}$ and
  $x_{out}$. $f^{(rot)}$ is a rotation depending on the distance from the mean
  $x_m$ of $x_{in}$ and $x_{out}$ in the previously selected two-dimensional
  plane.}
  \label{fig:f-rotation}
\end{figure}
\vspace{-6pt}

\begin{figure}[H]

  \begin{subfigure}[t]{0.48\textwidth}
  \includegraphics[width=.55\textwidth]{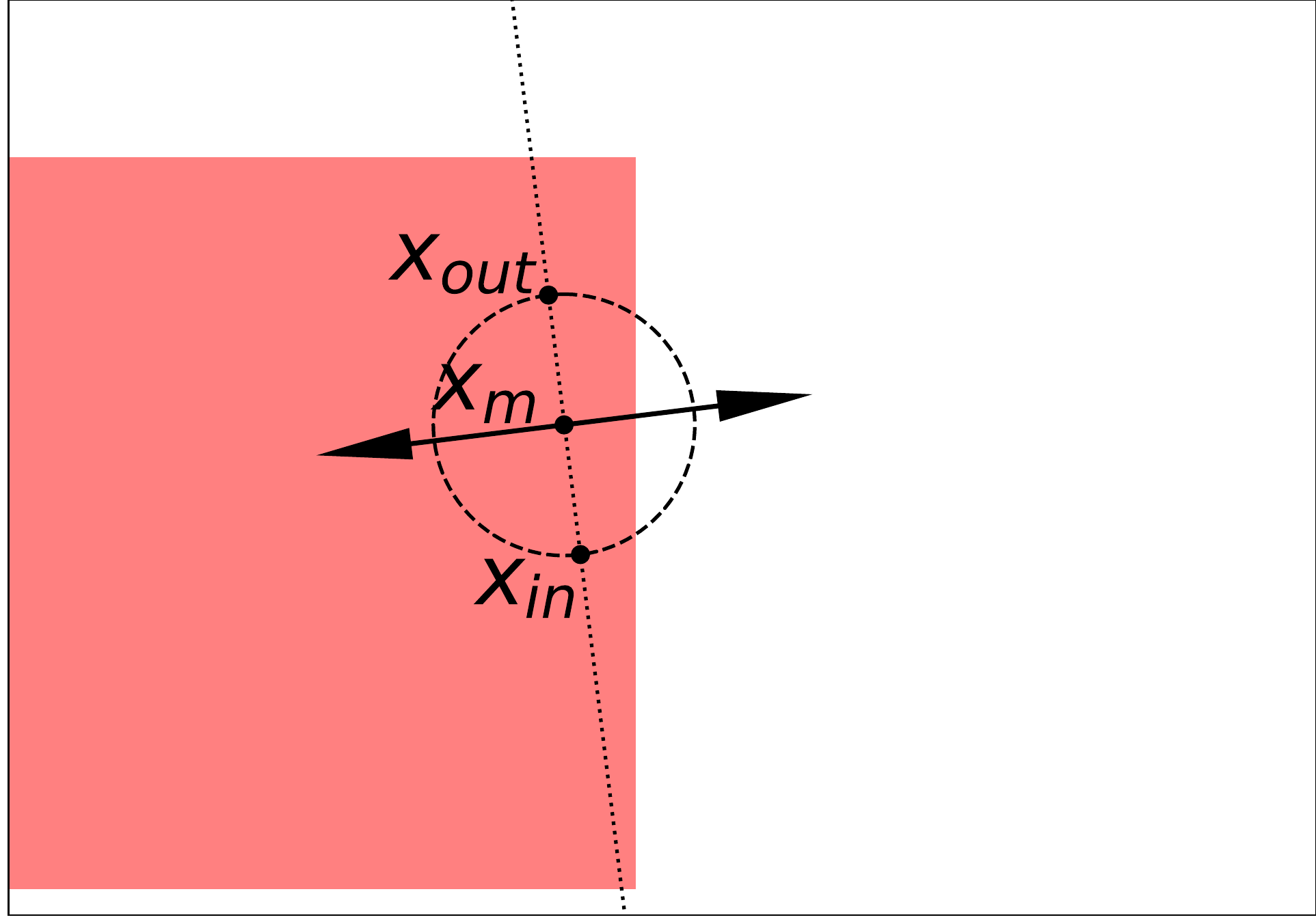}
  \caption{}
  \label{fig:ball-out}
  \end{subfigure}
  \hspace{-30pt}
  \begin{subfigure}[t]{0.48\textwidth}
  \includegraphics[width=.55\textwidth]{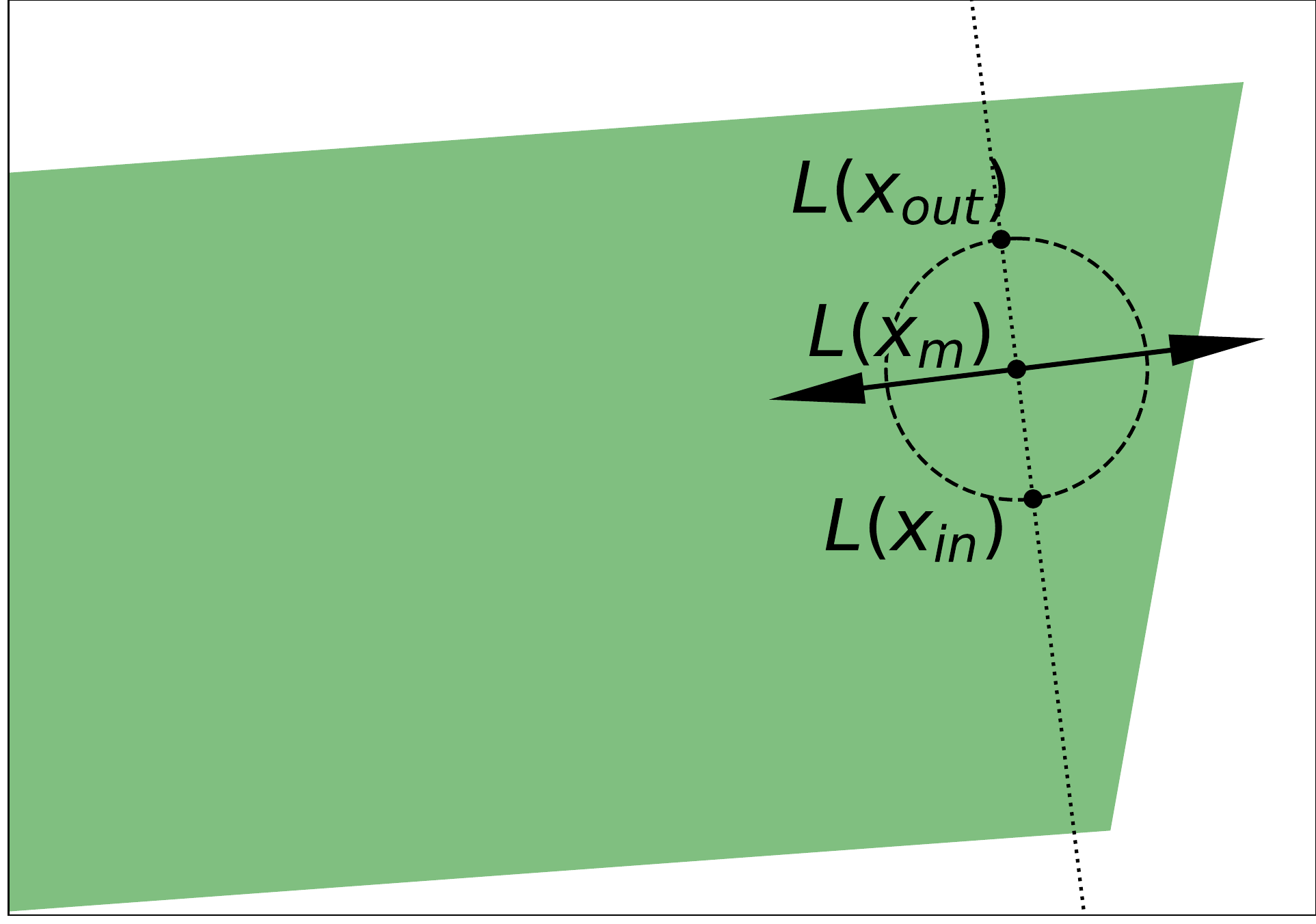}
  \caption{}
  \label{fig:ball-in}
  \end{subfigure}
  \vspace{6pt}
  \caption{We illustrate how, given $x_{in}$ and $x_{out}$ in a uniformly
  distributed hypercube $[0, 1]^D$, one can modify the space
  such that $f^{(rot)}$ shown in Figure \ref{fig:f-rotation} can be applied
  without modifying the distribution. (\textbf{a}) When taking two points $x_{in}$ and $x_{out}$ inside the hypercube
  $[0, 1]^D$, there is sometimes no
  $L_2$-ball centered in their mean $x_m$ containing both $x_{in}$ and
  $x_{out}$. (\textbf{b}) However, given $x_{in}$ and $x_{out}$, one can apply an invertible
  linear transformation $L$ such that there exists a $L_2$-ball centered in
  their new mean $L(x_m)$ containing both $L(x_{in})$ and $L(x_{out})$. If
  the distribution was uniform inside $[0, 1]^D$, then it is now
  also uniform inside $L\left([0, 1]^D\right)$.}
  \label{fig:put-ball-in}
\end{figure} 

Since the resulting distribution $p^*_{f(X)}$ is identical to the
original distribution $p^*_X$, their entropies are the same
$H\left(p^*_{f(X)}\right) = H\left(p^*_X\right)$. Hence, when $x_{in}$ and
$x_{out}$ are respectively an inlier and an outlier, whether in terms of
density
scoring or typicality, there exists a reparametrization of the problem
conserving the overall distribution while still exchanging their status of
inlier/outlier. We provide an example applied to a standard Gaussian
distribution in Figure \ref{fig:invariant-transform}.

\end{paracol}
\nointerlineskip
\begin{figure}[H]
\widefigure
  \centering
  \begin{subfigure}[t]{0.32\textwidth}
  \centering
  \begin{tikzpicture}
    \node (img)  {\includegraphics[width=.8\textwidth]{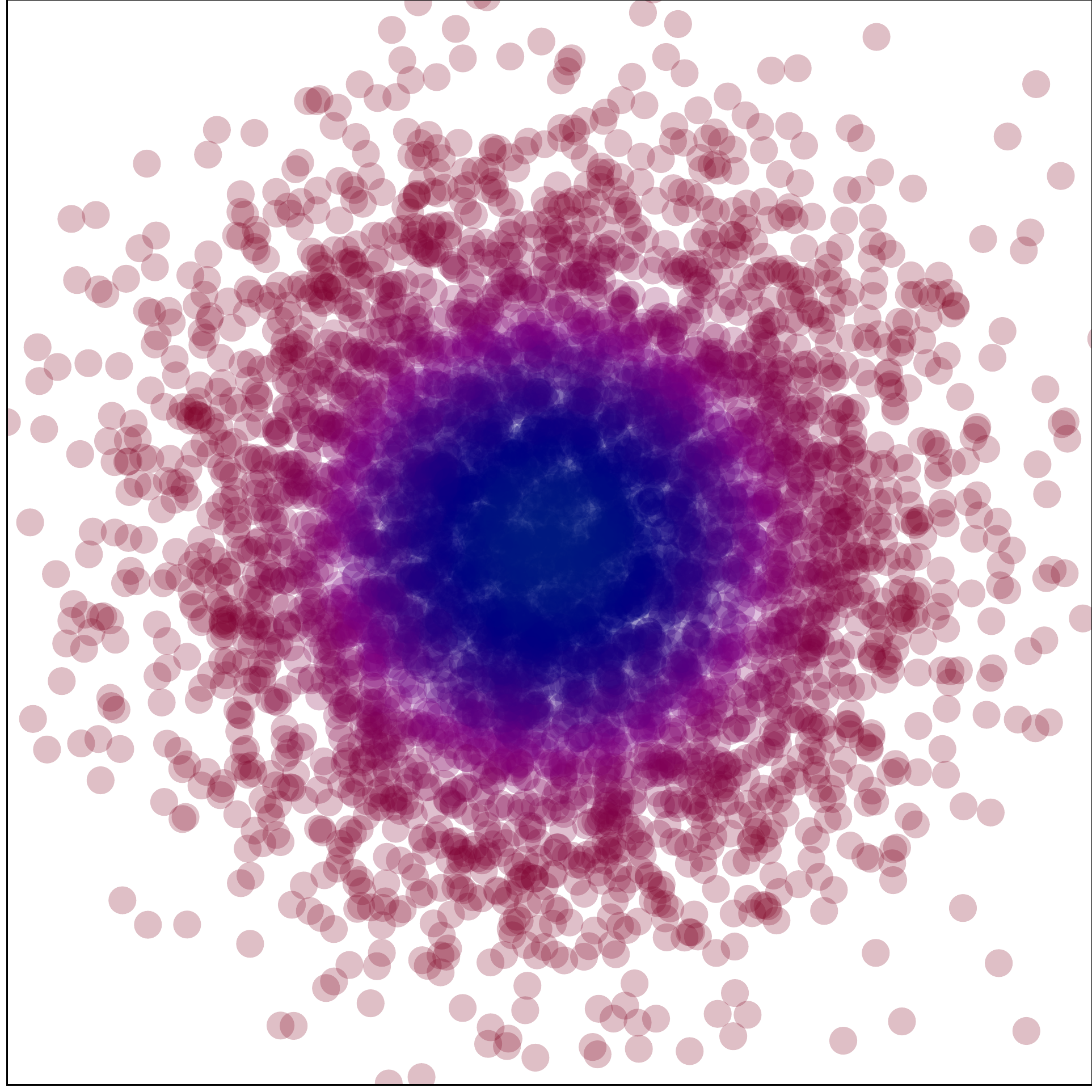}};
    \node[below of=img, node distance=2.9cm, yshift=0cm] {\small $x_1$};
    \node[left of=img, node distance=2.9cm, rotate=90, anchor=center] {\small $x_2$};
  \end{tikzpicture}
  \caption{}
  \label{fig:init-gaussian}
  \end{subfigure}
  \begin{subfigure}[t]{0.32\textwidth}
  \centering
  \begin{tikzpicture}
    \node (img)  {\includegraphics[width=.85\textwidth]{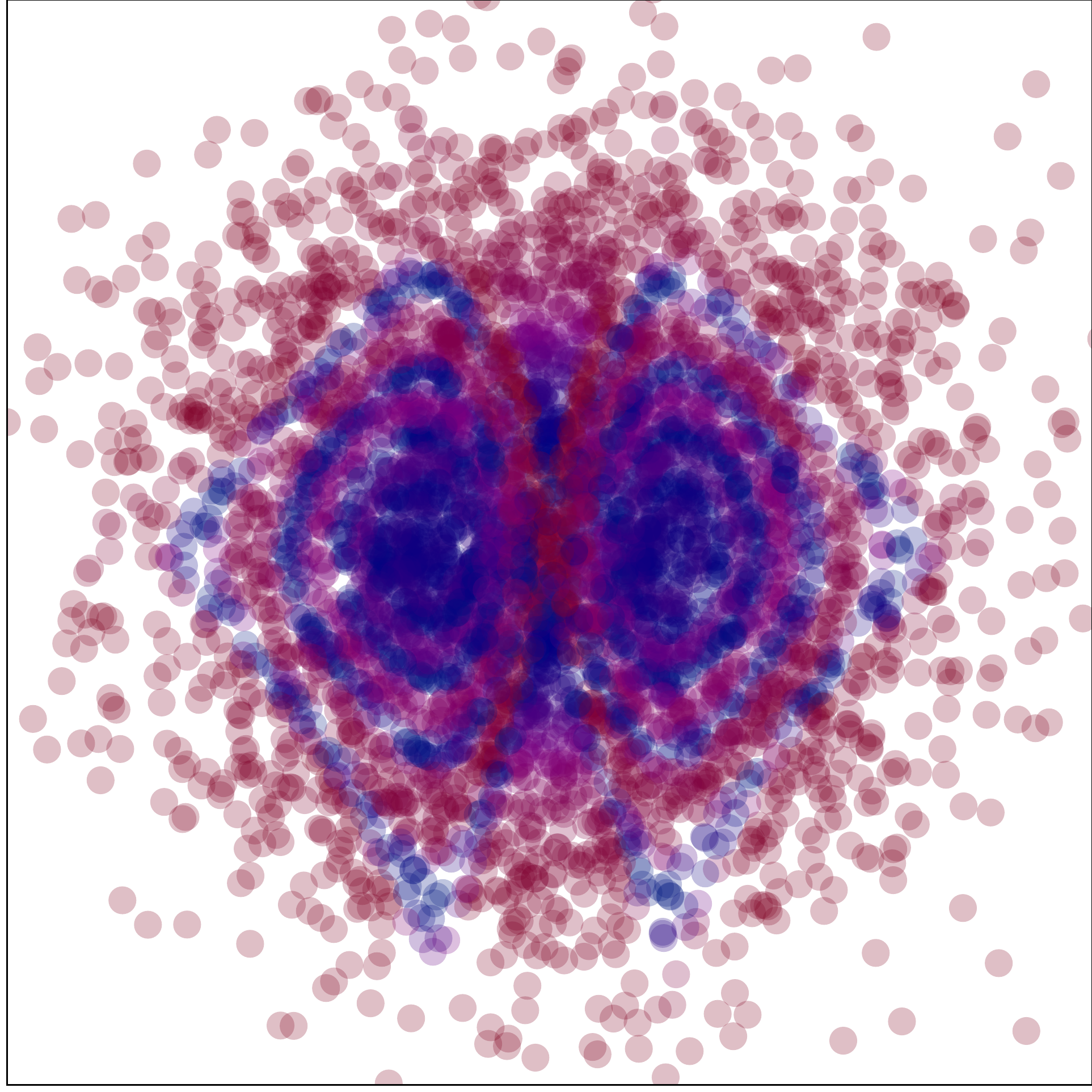}};
    \node[below of=img, node distance=2.9cm, yshift=0cm] {\small $\big(f(x)\big)_1$};
    \node[left of=img, node distance=2.9cm, rotate=90, anchor=center] {\small $\big(f(x)\big)_2$};
  \end{tikzpicture}
  \caption{}
  \label{fig:rotated-gaussian}
  \end{subfigure}
  \begin{subfigure}[t]{0.32\textwidth}
  \centering
  \begin{tikzpicture}
    \node (img)  {\includegraphics[width=.85\textwidth]{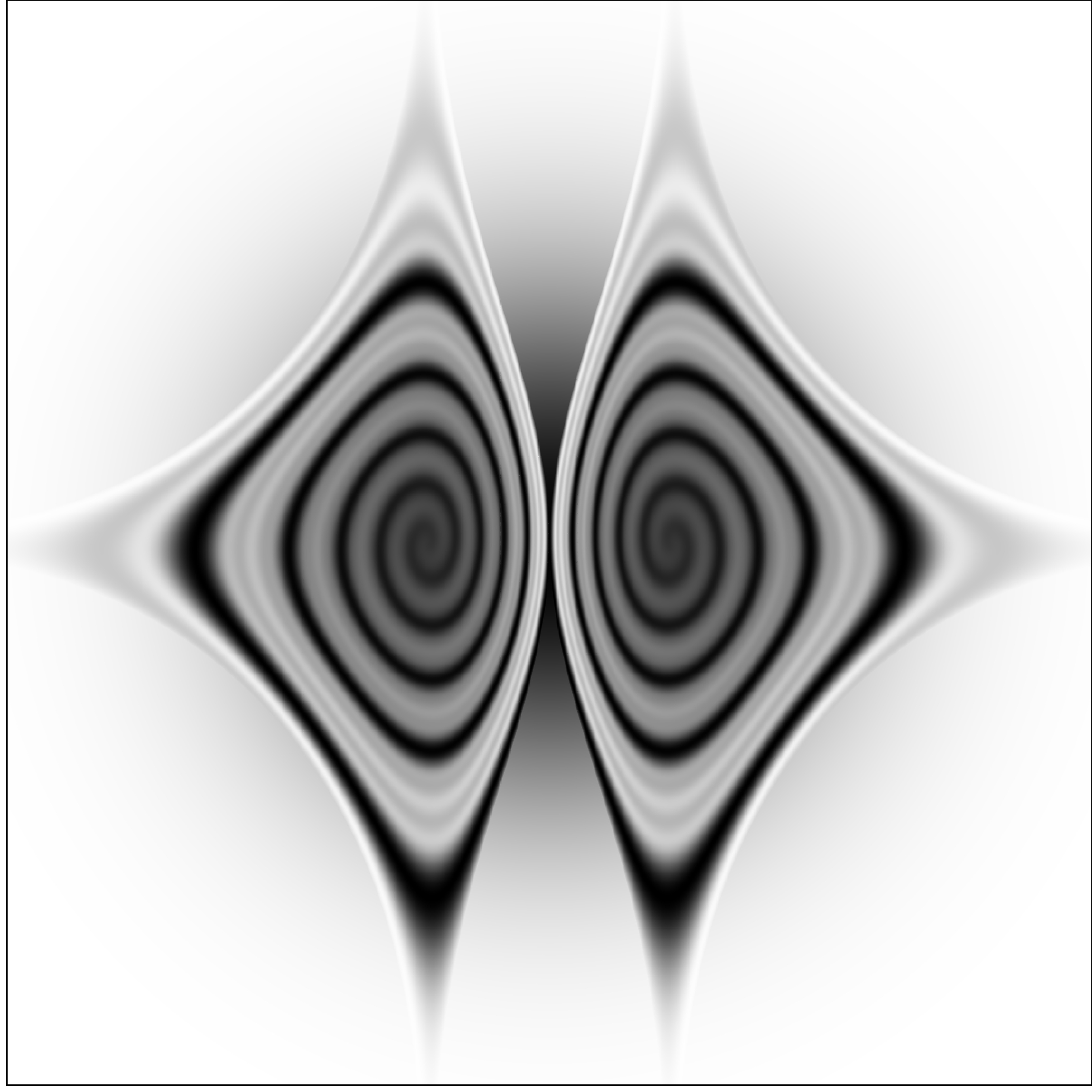}};
    \node[below of=img, node distance=2.9cm, yshift=0cm] {\small $\big(f(x)\big)_1$};
    \node[left of=img, node distance=2.9cm, rotate=90, anchor=center] {\small $\big(f(x)\big)_2$};
  \end{tikzpicture}
  \caption{}
  \label{fig:invar-pdf}
  \end{subfigure}
  \vspace{9pt}
  \caption{Application of the bijection from Figure \ref{fig:f-rotation} to a standard Gaussian distribution
  $\mathcal{N}(0, \mathbb{I}_2)$ leaving it an overall invariant. (\textbf{a}) Points sampled from $p^*_X = \mathcal{N}(0, \mathbb{I}_2)$. (\textbf{b}) Applying a bijection $f$ that preserves the distribution
  $p^*_{f(X)} = \mathcal{N}(0, \mathbb{I}_2)$ to the points in
  Figure \ref{fig:invariant-transform}a. (\textbf{c}) The original distribution $p^*_{X}$ with respect to the
  new coordinates $f(x)$: $p^*_{X} \circ f^{-1}$.}
  \label{fig:invariant-transform}
\end{figure}
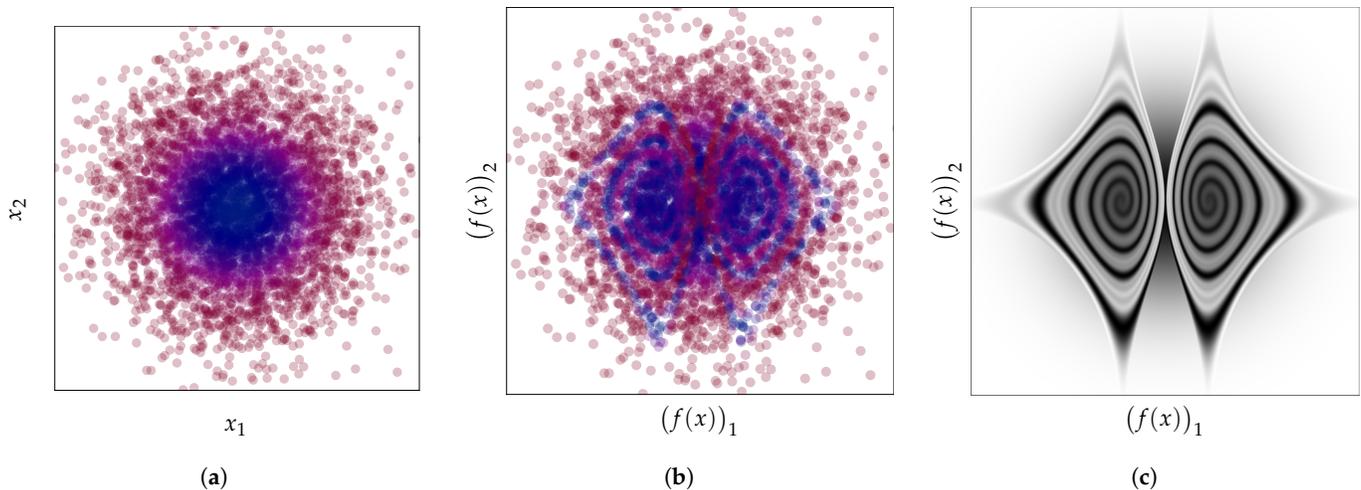 
\begin{paracol}{2}
\switchcolumn

This result is important from a
representation learning perspective and a complement to the general
non-identifiability result in several representation learning\linebreak approaches
\citep{hyvarinen1999nonlinear,locatello2019challenging}.
It means that learning a
representation with a predefined, well-known distribution and knowing the
true density $p^*_X$ are not sufficient conditions to control the
individual density of each point and accurately distinguish outliers from
inliers.

\subsection{Practical Consequences for Anomaly Detection}
We showed that the choice of representation can heavily influence the output of the anomaly detection methods described in Sections \ref{sec:densityscoring} and \ref{sec:typicality}.

\subsubsection{Learning a Representation by Applying Explicit Transformations f} 

Surprisingly, this problem can persist even when the learned representation is lower-dimensional, contains {\em only} the relevant information for the task, and is axis-aligned with semantic variables, since a reasoning similar to Section \ref{sec:arbitraryscoring} can be applied using axis-aligned bijections to tamper with densities. If a recent review \citep{ruff2021unifying} has highlighted the importance of the choice of representation in the context of low-level/high-level anomalies, our result goes further and shows that a problem still persists as even high-level information can be invertibly reparametrized to impose an arbitrary density-based ranking.
This leads us to believe that characterizing which representations are suitable for density-based methods (to conform with human expectations) cannot be answered in the absence of prior knowledge (see Section \ref{sec:arbitraryscoring}), e.g., on the distribution of anomalies. 

\subsubsection{Arbitrary Input Representation Result from Implicit Transformations f}

While (to our knowledge) input features are rarely designed or heavily tampered with to obfuscate density-based methods in practice, input features can often be the result of a system not fully understood end-to-end, that is of some \textit{implicit transformations $f$}, as to how they influence the task of anomaly detection. For instance, cameras used can be tuned to different tasks and the spectral response of film and image sensors has been tuned to maximize performance on the “Shirley Card” \citep{roth2009looking,buolamwini2018gender}.  Images can also go through processing techniques like high-dynamic range imaging~\citep{reinhard2010high} or arbitrary downsampling~as in \cite{torralba200880, krizhevsky2009learning, netzer2011reading}.

It is well-understood in {\em representation learning} \citep{bengio2013representation} that the default input features handed to the learning algorithm are rarely well-tuned to the task it tries to solve, e.g., euclidean distance rarely follows a notion of semantic distance, see \cite{theis2015note}. Figure \ref{fig:one-pixel1} provides an example where these methods fail in pixel space despite being endowed with a perfect density model. Details about its construction and analysis are provided below.

\begin{figure}[H]

  \centering\includegraphics[width=.55\textwidth]{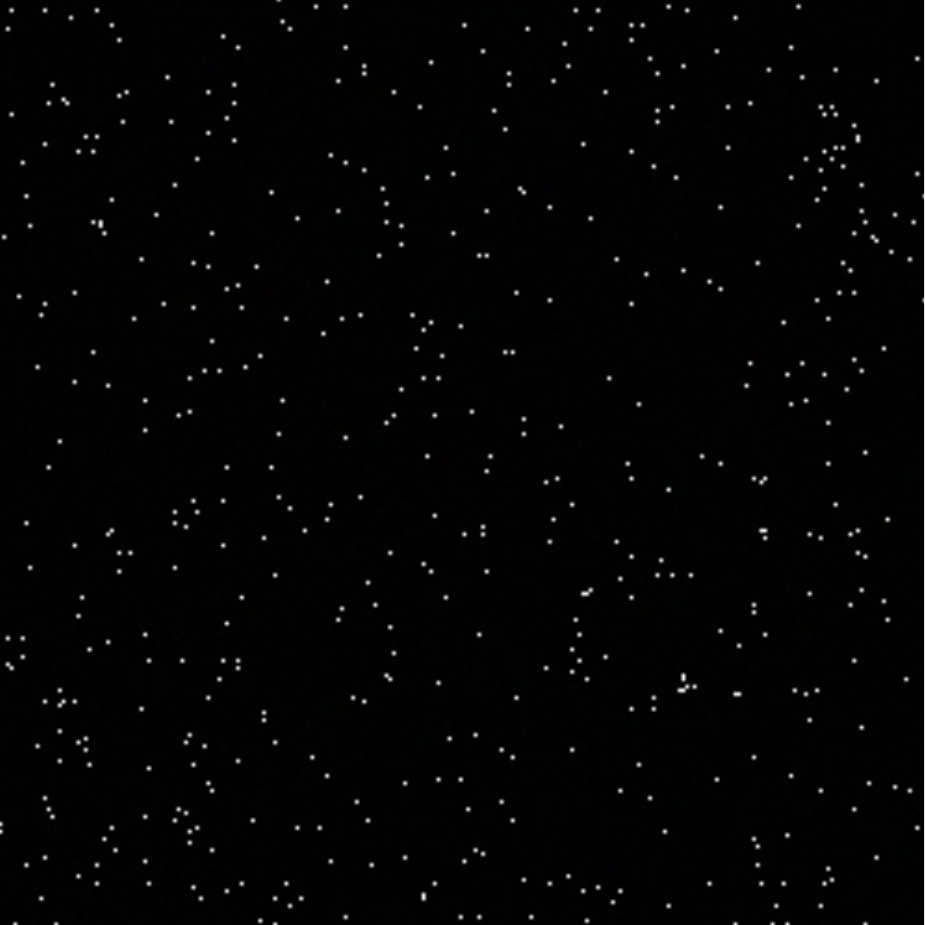}
  \vspace{6pt}
  \caption{We generated $5^6$ pixels according to the procedure described and concatenated them in a single $125 \times 125$ RGB bitmap image for an easier visualization. While, visual intuition would suggest that white pixels are the outliers in this figure, density-based definitions of anomalies described Section \ref{sec:densityscoring} (density scoring) and Section \ref{sec:typicality} (typicality) would consider a specific dark shade of gray to be the outlier.}
  \label{fig:one-pixel1}
\end{figure}

We generate $5^6$ individual pixels as three-dimensional vectors according to a distribution built as follows:
let $p_{w} = \mathcal{U}([255, 256]^3)$ (corresponding to the color white),
$p_{b} = \mathcal{U}([0, 10]^3)$ (corresponding to shades of black),
and $p_{out} = \mathcal{U}([10, 11]^3)$ (corresponding to a dark shade of grey) be distributions with disjoint supports. We consider pixels following the distribution
\begin{align*}
p_X(x) = &\beta \cdot p_{out}(x) \\
+ (1 - &\beta) \big(\alpha \cdot p_{w}(x) + (1 - \alpha) \cdot p_{b}(x)\big),
\end{align*}
where $\alpha = 1001^{-3}$ and $\beta = 10^{-4}$.
Once generated, we concatenate these pixels in a $125 \times 125$ RGB bitmap image in Figure \ref{fig:one-pixel1} for a more convenient visualization.

Visually, a common intuition would be to consider white pixels to be the anomalies in this figure.
However, following a construction similar to Section \ref{sec:arbitraryscoring},
the final densities corresponding to pixels from
$p_{w}$ (equal to $\alpha (1 - \beta)$)
and $p_{b}$ (equal to $(1 - \alpha) (1 - \beta) 10^{-3}$)
are equal to $1001^{-3} (1 - 10^{-4}) \approx 10^{-3}$,
and the final density corresponding to pixels from
$p_{out}$ (equal to $\beta$) is $10^{-4}$.
Therefore, none of the methods presented in
Section \ref{sec:densityscoring} (density scoring) and Section \ref{sec:typicality} (one-sample typicality) would consider
the white pixels (in $[255, 256]^3$) as outliers. They would only classify the pixels of a particular dark shade of gray in $[10, 11]^3$ as outliers.

Given the considerable influence, the choice of input representation has on the output of even the true data density function $p^*_X$, one should question the strong but understated assumption behind current practices that (density-based) anomaly detection methods applied on \textit{default input representations} decontextualized from their design process \cite{raji2020everything}, \textit{representations orthogonally learned from the task}, or \textit{even obtained by filtering noise variables} (non-semantic) ought to result in proper outlier classification.

\section{Promising Avenues for Unsupervised Density-Based Anomaly Detection}
\label{sec:prior}

While anomaly detection can be an
ill-posed problem~as mentioned in \cite{choi2018waic, nalisnick2019detecting,
morningstar2020density} without \textit{prior knowledge}, several approaches are more promising by making this prior knowledge more explicit.
We highlighted the strong dependence of density-based anomaly detection methods on a choice of representation, which needs to be justified as it is crucial to the success of the approach.
This was proven by using the change of variables formula, which describes how the density function varies with respect to a reparametrization. If we consider the fundamental definition of a density as a Radon-Nikodym derivative $p^*_X = \frac{d P^*_X}{d \mu_X}$ with respect to a base measure (here the Lebesgue measure $\mu_X$ in $\mathcal{X}$), we notice that this variation stems from a change of “denominator”: the Lebesgue measure corresponding to $\mathcal{X}$ is different to the one corresponding to another space $\mathcal{Z}$ (the Jacobian determinant accounting for this mismatch $\mu_X \circ X \neq \mu_Z \circ Z$). 

 A way to incorporate more transparently the choice of representation is to consider a similar fraction. For example, {\em density ratio methods}~\citep{griffiths2007mere} score points using a ratio $p^*_X / p_{BG}$ between two densities. The task is then to figure out whether a point comes from a regular source (the foreground distribution in the numerator) or an anomalous source (the background distribution in the denominator). The concurrent work \citep{zhang2021understanding} also draws a similar conclusion showing that no test can distinguish between a given source distribution and an unspecified outlier distribution better than random chance. In \citet{bishop1994novelty}, the density scoring method has been interpreted as a density ratio method with a default uniform density function. More refined methods can be used as a background distribution, e.g., $p^*_X$ convolved with a noise distribution~\citep{ren2019likelihood}, the implicit distribution of a compressor~\citep{Serra2020Input}, or a mixture including $p^*_X$ as a component, i.e., a “superset”, see \cite{schirrmeister2020understanding}. In addition to being more transparent with respect to its underlying assumptions, density ratio methods are invariant to invertible reparametrization.

While appealing in their property, density ratio methods still require the explicit definition of a background distribution, an explicit guess on how the anomalies should be distributed. It is actually possible in some cases to be more intentional in the definition of this denominator. For example, for exploration in reinforcement learning, \mbox{Houthooft et al.  \cite{houthooft2016vime}} and Bellemare et al. \cite{bellemare2016unifying} use an (invertible) reparametrization-invariant proxy for potential information gain.

\section{Discussion and Limitations}
\label{sec:discussion}
 \looseness=-1 We discussed the ill-defined (and arguably subjective) notion of {\em outlier} or {\em anomaly}, which several works attempted to characterize through a seemingly clearer notion of probability density used in the density scoring and typicality test methods.
We show in this paper that an undesirable degree of freedom persists in how density functions can be manipulated by an arbitrary choice of representation, rarely set to fit the task.
We consider that the lack of attention paid to this crucial element has undermined the foundations of these off-the-shelf methods, potentially providing a simpler explanation to their empirical failures~studied in \cite{choi2018waic, nalisnick2018deep, hendrycks2018deep, just2019deep, Fetaya2020Understanding, kirichenko2020normalizing} as a discrepancy with unstated prior assumptions.

We conclude that being more intentional about integrating prior knowledge explicitly in density-based anomaly detection algorithms then becomes essential to their success.

Although a similar issue persists in practice for {\em discrete spaces}  as noted in \cite{dieleman2020typicality}, where outputs with highest probability are atypical, 
the same reparametrization trick used throughout this paper to formalize this issue for continuous inputs is not directly applicable for discrete input spaces.
However, similar adversarial constructions can be made in an analogous way: semantically close inputs can be considered distinct or identical depending on arbitrary choices of discretization/categorization~\citep{hanna2020towards}, resulting in different probability values. Arbitrary choices of discretization include tokenization, lemmatization, or encoding~see \cite{wang2020neural} for language modeling but also choice of language~\citep{de2019does}. Figure \ref{fig:one-pixel1} provides a similar construction in discrete pixel space.

Similarly, while approaches involving {\em probability masses} are unaffected by invertible reparametrizations, they explicitly rely on a deliberate choice in partitioning the input space, which is why we consider such approaches coherent with a more explicit incorporation of prior knowledge.

We make in the paper the assumption that the data distribution density {\em $p^*_X$ is strictly positive} everywhere in the set of possible instances $\mathcal{X}$ since in practice deep density models spread probability over all the input space. Arguably, an instance occurring outside the support of the data distribution would be considered an anomaly.
An example would be CIFAR-10 and SVHN, which can be assumed to be disjoint. However, considering even the slightest Gaussian noise on either data distribution is sufficient to have \textit{non-disjoint supports} as it makes the densities non-zero everywhere in the pixel space. Since Section \ref{sec:typicality} highlighted a failure of our geometrical intuition of density through the Gaussian Annulus theorem, we advocate for some skepticism on the assumption that these data distributions ought to be completely disjoint.
In the general case, it is unknown whether anomalies lie outside of the distribution support and not uncommon to consider the probability of an anomaly happening to be non-zero with respect to the data distribution (i.e., $P^*_X(\mathcal{X}_{out}) > 0$), which is coherent with this strict positivity assumption. On the contrary, the concurrent work \citep{zhang2021understanding} chooses to assume a disjoint support for the inlier and outlier distributions, leading them to conclude that the model misestimation is the source of the observations made by Nalisnick et al.  \cite{nalisnick2018deep}.

\section{Broader Impact}
Anomaly detection is commonly proposed as a fundamental element to safely deploy machine learning models in the real world. Its applications range from medical diagnostics and autonomous driving to cyber security and financial fraud detection.The use of such models on outlier points can result in dangerous behaviors but also discriminatory outcomes. Our paper aims at questioning current density-based anomaly detection methods, which is essential to mitigate the risks associated with their use in the real-world. 

More broadly, our study also leads to reconsider the role of density as a standalone quantity and practices built around it, e.g., temperature sampling~\citep{graves2013generating, kingma2018glow, pmlr-v80-parmar18a} and evaluating density models on anomaly detection, e.g., as in \cite{du2019implicit, grathwohl2019your, kirichenko2020normalizing,
liu2020hybrid}. 

Finally, a common opinion in machine learning~\citep{kurenkov2020lessons} has been that, given enough data and capacity, machine learning bias generally has a vanishing influence over the resulting bias in the learned solution.
On the contrary, scale can obfuscate~\citep{raji2020everything} misspecifications in the task and/or data collection design~\citep{birhane2021large, paullada2020data}.
Here, we focused on how misspecifications in the algorithm design for anomaly detection can result in gross failure even in the ideal theoretical settings of infinite data and capacity.

However, this study provides a constructive proof in Section \ref{sec:arbitraryscoring} that bad actors can use to arbitrarily manipulate the results of currently used anomaly detection algorithms, without modifying a learned model $p^{(\theta)}_X$. This opens the door to potential negative impacts if unreasonable trust in these methods are maintained in practice. 

\vspace{6pt} 

\authorcontributions{Conceptualization, L.D.; methodology, L.D.; formal analysis, C.L.L. and L.D.; investigation, C.L.L. and L.D.; writing---original draft preparation, C.L.L. and L.D.; writing---review and editing, C.L.L. and L.D.; visualization, L.D.; supervision, L.D. All authors have read and agreed to the published version of the manuscript.}

\funding{This research received no external funding.}

\institutionalreview{Not applicable.}

\informedconsent{Not applicable.}

\dataavailability{Not applicable.}

\acknowledgments{
The authors would like to thank
Kyle Kastner,
Johanna Hansen,
Ben Poole,
Arthur Gretton,
Durk Kingma,
Samy Bengio,
Jascha Sohl-Dickstein,
Adam Foster,
Polina Kirichenko,
Pavel Izmailov,
Ross Goroshin,
Hugo Larochelle,
J\"{o}rn-Henrik Jacobsen,
and Kyunghyun Cho
for initial discussions for this paper. We also thank
Eric Jang,
Alex Alemi,
Jishnu Mukhoti,
Jannik Kossen,
Sebastian Schmon,
Francisco Ruiz,
David Duvenaud,
Luke Vilnis,
and the anonymous reviewers for useful feedback on this paper. We would also like to thank
the Python \mbox{community \citep{van1995python,oliphant2007python}} for developing
tools that enabled this work, including
{NumPy}~\citep{oliphant2006guide,walt2011numpy, harris2020array},
{SciPy}~\citep{jones2001scipy}, and
{Matplotlib}~\citep{hunter2007matplotlib}.
}

\conflictsofinterest{The authors declare no conflict of interest.}

\appendixtitles{yes} %
\appendixstart
\appendix
\section{Proof of Proposition \ref{prop:invariant}}
\label{app:proof}
\proptwo*
\begin{proof}
Our proof will rely on the following non-rigid rotation $f^{(rot)}$. Working in a hyperspherical
coordinate system consisting of a radial coordinate $r>0$ and $(D-1)$ angular coordinates $(\phi_i)_{i < D}$,
\begin{align*}
\forall d<D,~x_d &= r\left(\prod_{i=1}^{d-1}{\sin(\phi_i)}\right)\cos(\phi_d) \\
x_D &= r\left(\prod_{i=1}^{D-2}{\sin(\phi_i)}\right)\sin(\phi_{D-1}),
\end{align*}
where for all $i\in \{1, 2, ..., D-2\},$ $\phi_i \in [0, \pi)$ and $\phi_{D-1} \in [0, 2\pi)$, 
given $r_{max} > r_0 > 0$, we define the continuous mapping $f^{(rot)}$ as:
\begin{align*}
f^{(rot)}\big((r, \phi_1, &\dots, \phi_{D-2}, \phi_{D-1})\big) \\
=\Big(\Big.r, \phi_1, &\dots, \phi_{D-2}, \\
&\phi_{D-1} + \pi \frac{(r_{max} - r)_+}{r_{max} - r_0} [\text{mod}~2\pi]\Big.\Big) .
\end{align*}
where $(\cdot)_+ = \max(\cdot, 0)$.
This mapping only affects points inside $\mathcal{B}_2(0, r_{max})$,
and exchanges two points corresponding
to $(r_0, \phi_1, \dots, \phi_{D-2}, \phi_{D-1})$ and
$(r_0, \phi_1, \dots, \phi_{D-2}, \phi_{D-1} + \pi)$ in a continous way (see
Figure \ref{fig:f-rotation}). Since the Jacobian determinant of the
hyperspherical coordinates transformation is not a function of $\phi_{D-1}$,
$f^{(rot)}$ is volume-preserving in \mbox{cartesian coordinates}.

Let $f^{(KR)}$ be a Knothe-Rosenblatt rearrangement of $p^*_X$,
$f^{(KR)}(X)$ is uniformly distributed in $[0, 1]^D$.
Let $z^{(in)} = f^{(KR)}\left(x^{(in)}\right)$
and $z^{(out)} = f^{(KR)}\left(x^{(out)}\right)$.
Since $f^{(KR)}$ is continuous,
$z^{(in)}, z^{(out)}$
are in the interior $(0, 1)^D$.
Therefore, there is an $\epsilon > 0$ such that the $L_2$-balls
$\mathcal{B}_2\left(z^{(in)}, \epsilon\right)$ and
$\mathcal{B}_2\left(z^{(out)}, \epsilon\right)$
are inside $(0, 1)^D$.
Since $(0, 1)^D$ is convex, so is their convex hull.

Let $r_0 = \frac{1}{2}\norm{z^{(in)} - z^{(out)}}_2$
and $r_{max} = r_0 + \epsilon$.
Given $z \in (0, 1)^D$, we write $ z_\parallel$ and $z_\perp$ to denote its parallel and orthogonal
components with respect to $\left(z^{(in)} - z^{(out)}\right)$.
We consider the linear bijection $L$ defined by
\[L(z) = z_\parallel + \epsilon^{-1} r_{max} z_\perp.\]

Let $f^{(z)} = L \circ f^{(KR)}$. Since $L$ is a linear function
(i.e., with constant Jacobian), $f^{(z)}(X)$ is uniformly distributed inside
$L\left([0, 1]^D\right)$.
If $z^{(m)}$ is the mean of
$z^{(in)}$ and $z^{(out)}$, then
$f^{(z)}(\mathcal{X})$ contains
$\mathcal{B}_2\left(L\left(z^{(m)}\right), r_{max}\right)$
(see Figure \ref{fig:put-ball-in}).
We can then apply the non-rigid rotation $f^{(rot)}$ defined earlier,
centered on $L\left(z^{(m)}\right)$ to exchange
$L\left(z^{(in)}\right)$ and
$L\left(z^{(out)}\right)$ while maintaining this uniform distribution.

We can then apply the bijection $\left(f^{(z)}\right)^{-1}$ to obtain the
invertible map $f = \left(f^{(z)}\right)^{-1} \circ f^{(rot)} \circ f^{(z)}$
such that  $p^*_{f(X)} = f^*_X$,
$p^*_{f(X)}\left(f\left(x^{(in)}\right)\right) = p^*_X\left(x^{(out)}\right)$,
and\linebreak
$p^*_{f(X)}\left(f\left(x^{(out)}\right)\right) = p^*_X\left(x^{(in)}\right)$.
\end{proof}

\end{paracol}
\reftitle{References}

\end{document}